\documentclass[10pt,twocolumn,letterpaper]{article}

\usepackage[pagenumbers]{cvpr} % To force page numbers, e.g. for an arXiv version

\usepackage{graphicx}
\usepackage{amsmath}
\usepackage{amssymb}
\usepackage{booktabs}

\usepackage{multirow}
\usepackage{tabularx}
\usepackage{enumitem}

\usepackage{longtable} % for multipage VTAB table
\usepackage{xcolor,colortbl}

\usepackage{soul}  % For text highlighting.
\usepackage{setspace}  % Line spacing

\usepackage{listings}
\usepackage[pagebackref,breaklinks,colorlinks]{hyperref}

\usepackage[capitalize]{cleveref}
\crefname{section}{Sec.}{Secs.}
\Crefname{section}{Section}{Sections}
\Crefname{table}{Table}{Tables}
\crefname{table}{Tab.}{Tabs.}

\def \flame {\raisebox{-.1\height}{\includegraphics[height=0.7\baselineskip]{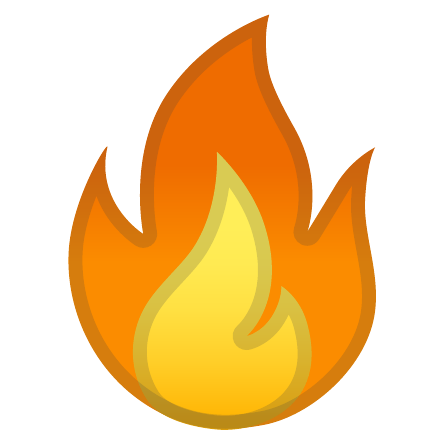}}\xspace}
\def \litf {LiT\flame{}\xspace}
\def \lit {LiT\xspace}
\def \Lu {{\tt Lu}\xspace}
\def \LU {{\tt LU}\xspace}
\def \Uu {{\tt Uu}\xspace}
\def \UU {{\tt UU}\xspace}
\def \uu {{\tt uu}\xspace}

\definecolor{light_gray}{rgb}{0.8, 0.8, 0.8}
\DeclareRobustCommand{\hlgrey}[1]{{\sethlcolor{light_gray}\hl{#1}}}
\newcommand{\clsfmt}[1]{\hlgrey{\textit{#1}}}
\def \cls {\clsfmt{CLASS}}

\renewcommand*{\eg}{e.g.\@\xspace}
\renewcommand*{\ie}{i.e.\@\xspace}
 
\makeatletter
\renewcommand{\paragraph}{%
  \@startsection{paragraph}{4}%
  {\z@}{1.75ex \@plus 1ex \@minus .2ex}{-1em}%
  {\normalfont\normalsize\bfseries}%
}
\makeatother

\def\objsota{82.5}

\def\cvprPaperID{10338} % *** Enter the CVPR Paper ID here
\def\confName{CVPR}
\def\confYear{2022}

\begin{document}
\lstMakeShortInline`

\title{\litf{}: Zero-Shot Transfer with Locked-image text Tuning}

\newcommand{\authsep}{  }

\author{\normalsize{Xiaohua Zhai$^{\star\dagger}$ \authsep Xiao Wang$^{\star}$ \authsep Basil Mustafa$^{\star}$ \authsep Andreas Steiner$^{\star}$ \authsep Daniel Keysers \authsep Alexander Kolesnikov \authsep Lucas Beyer$^{\star\dagger}$}\\
Google Research, Brain Team, Zürich}
\maketitle
{\let\thefootnote\relax\footnote{
{$^{\star}$equal technical contribution, $^{\dagger}$equal advising}}}

\begin{abstract}

This paper presents \emph{contrastive-tuning}, a simple method employing contrastive training to align image and text models while still taking advantage of their pre-training. In our empirical study we find that locked pre-trained image models with unlocked text models work best. We call this instance of contrastive-tuning “Locked-image Tuning” (\lit{}), which just teaches a text model to read out good representations from a pre-trained image model for new tasks. A \lit{} model gains the capability of zero-shot transfer to new vision tasks, such as image classification or retrieval. The proposed \lit{} is widely applicable; it works reliably with multiple pre-training methods (supervised and unsupervised) and across diverse architectures (ResNet, Vision Transformers and MLP-Mixer) using three different image-text datasets. With the transformer-based pre-trained ViT-g/14 model, the \lit{} model achieves 85.2\% zero-shot transfer accuracy on the ImageNet test set, and \objsota{}\% on the challenging out-of-distribution ObjectNet test set. 
\end{abstract}

% ===============================================
% ===============================================

\section{Introduction}\label{sec:intro}

\begin{figure}[t]
    \centering
    \includegraphics[width=\linewidth]{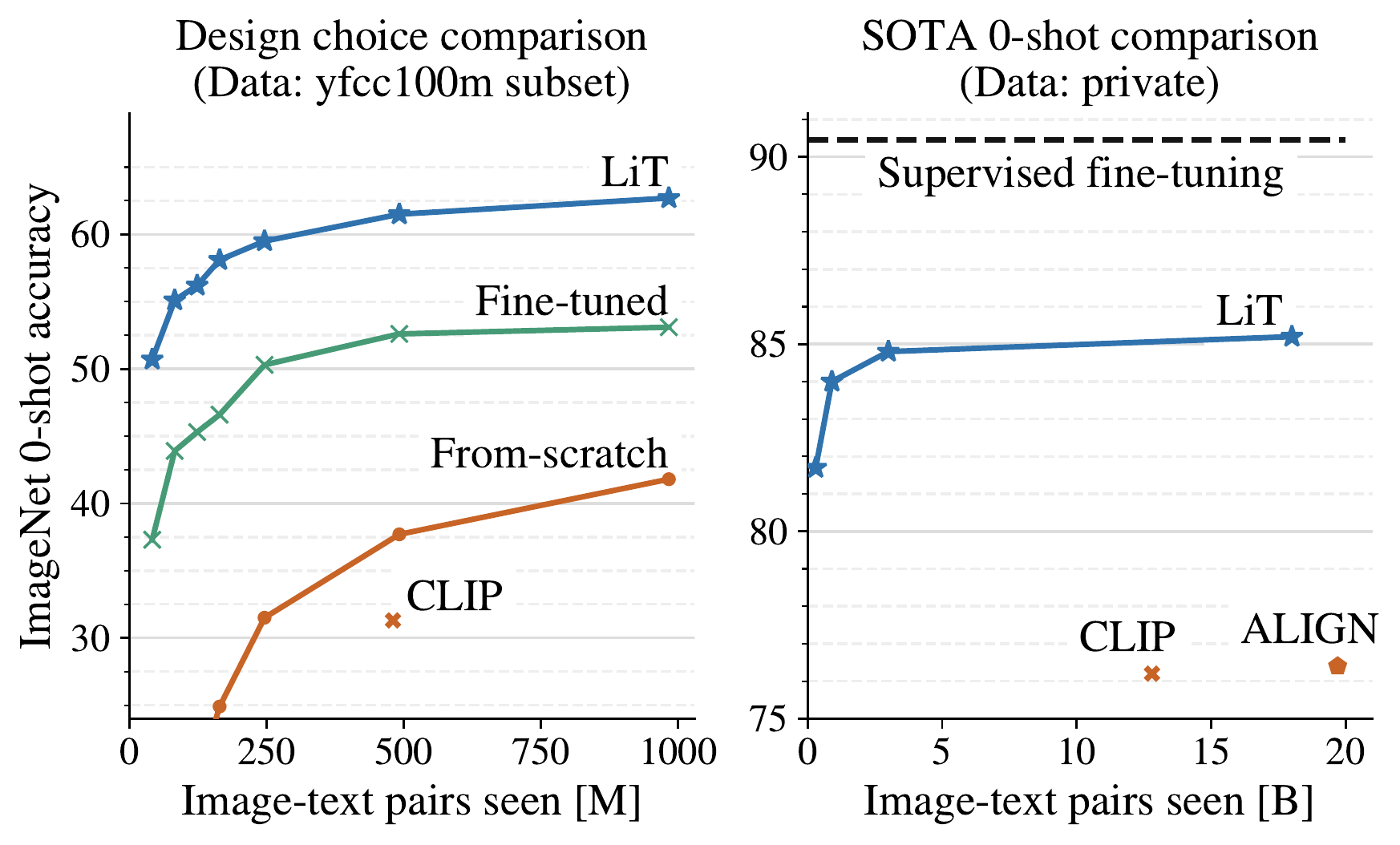}
    \caption{Comparison to the previous SOTA methods. \textbf{Left}: results on public YFCC100m subset, with from-scratch, fine-tuned from a pre-trained image model, and LiT with a pre-trained image model. The proposed LiT improves over 30\% ImageNet zero-shot transfer accuracy on YFCC100m subset.
    \textbf{Right}: results on privately gathered data, LiT halves the gap between previous from-scratch methods CLIP~\cite{clip}, ALIGN~\cite{align} and supervised fine-tuning~\cite{vitg,coatnet}. 
    }
    \label{fig:teaser_curve}
\end{figure}

Transfer learning~\cite{transfer_learning_survey} has been a successful paradigm in computer vision~\cite{imagenet_transfer_better,bit,instagram_resnext}.
Zero-shot learning~\cite{zero_data_learning,visual_attribute,zsl_good_bad_ugly} is an alternative approach aiming to develop models that can handle a new task without task-specific data or adaptation protocols.
Recently it was demonstrated that web-sourced paired image-text data can be used to pre-train strong models for zero-shot transfer~\cite{clip,align}. 
Zero-shot \textit{transfer} differs from classical zero-shot learning in that the transfer setup may see relevant supervised information during pre-training; it is zero-shot insofar as no supervised examples are used during the transfer protocol.
\mbox{GPT-3}~\cite{gpt3} explored a similar zero-shot transfer setup using model prompting via natural language. 

In \cite{clip,align} authors propose a contrastive learning frame\-work where an image model (or image tower) is trained simultaneously with a text model (or text tower). 
Both towers are trained to minimize a contrastive loss, which encourages representations of paired images and texts to be similar, and %
%, conversely, 
representations of non-paired images and texts to be dissimilar. 
At test time, the resulting model can be used for zero-shot image classification by comparing the image embedding with embeddings of textual class descriptions.

In this paper, we adopt a contrastive learning framework and propose a more data- and compute-efficient strategy named \textit{contrastive-tuning}. 
The key idea is to tune the text tower using image-text data, while using a pre-trained, strong image model as the image tower. 
During training, both towers' weights can be locked or unlocked, leading to different design choices that are illustrated in Figure~\ref{fig:teaser_design}.
Specifically, we find that locking the image tower works best, as shown in Figure~\ref{fig:teaser_curve}. 
We call this specific instance of contrastive-tuning “Locked-image Tuning” (\lit{}),
which just teaches a text model to read out suitable representations from a pre-trained image model. % for new tasks.
\lit{} achieves better results compared with the from-scratch CLIP~\cite{clip} or ALIGN~\cite{align} models.
With the pre-trained model ViT-g/14~\cite{vitg}, \lit{} achieves 85.2\% zero-shot transfer accuracy on ImageNet, halving the gap between previous best zero-shot transfer results~\cite{clip,align} and supervised fine-tuning results~\cite{vitg,coatnet}. The best \lit{} model also sets new state-of-the-art on several out-of-distribution (OOD) ImageNet test variants, compared to previous supervised and unsupervised methods.
For example, it achieves \objsota{}\% accuracy on the challenging ObjectNet test set~\cite{objectnet}, outperforming the previous state-of-the-art method~\cite{clip} by 10.2\%. 

We believe the reason that \lit{} works well lies in its decoupling of data sources and techniques for learning image descriptors and vision-language alignment. Image-text data can be great for learning correspondences between natural language and the visual world, but, at the same time, it may not be precise and clean enough to result in state-of-the-art image descriptors. In this paper we carefully investigate this hypothesis and support it with empirical evidence.

The proposed \lit{} works with both supervised and self-supervised pre-trained models.
We verify \lit{} across three image-text datasets, with Vision Transformer~\cite{vit}, ResNet~\cite{bit}, and MLP-Mixer~\cite{mixer} architectures. 
We also show that with a self-supervised pre-trained model, \ie DINO~\cite{dino} or MoCo-v3~\cite{mocov3}, \lit{} achieves better performance compared to from-scratch contrastive-learning. 

Another contribution of this paper is the proposed recipe for high-performance zero-shot models that can be trained using only modest computational resources and public datasets.
By re-using already pre-trained models (\eg publicly released in the literature), the computational resources used to train the image models can be amortized.
Furthermore, we explore publicly available datasets such as YFCC100m~\cite{yfcc100m} and CC12M~\cite{cc12m}.
Combined with the computational efficiency, we hope to facilitate contributions from a wider audience to research in zero-shot transfer. \footnote{Public LiT models available at \url{https://github.com/google-research/vision_transformer\#lit-models}. We provide pre-training code in the {\tt big\_vision} codebase~\cite{big_vision}.}

% ===============================================
% ===============================================

\section{Related work}\label{sec:rw}

This work is closely related to a vast amount of literature on \textit{transfer learning} in vision~\cite{transfer_learning_survey,survey2018}.
The main idea of transfer learning is to leverage already pre-trained models to solve a new task better and faster, as opposed to less efficient training from-scratch. 
This paradigm is usually implemented as a two-step procedure: (1)~pre-train (once) an initial model on a large dataset of images that are (weakly)-labeled or using self-supervised losses and (2)~fine-tune the pre-trained model for a task of interest using supervised data.
In the context of modern deep learning, many earlier works~\cite{decaf,off-the-shelf,imagenet_transfer_better,bit} used supervised pre-training to learn transferrable feature representations, with the Vision Transformer revisiting and improving this approach~\cite{vit,vitg}.
It was shown that scaling up model and dataset sizes simultaneously leads to dramatic improvements in transfer effectiveness~\cite{bit,vit,vitg} and robustness~\cite{robustness_transferability_cnn}.
Crucially, large pre-trained models exhibit outstanding capabilities in learning in the low-data (few-shot) regime~\cite{bit,vit,ting_ss}.

Still, collecting task-specific data and fine-tuning large pre-trained models remains time-consuming and potentially costly in many realistic scenarios. 
\textit{Zero-shot transfer} is an alternative paradigm that sidesteps the fine-tuning stage entirely and performs classification solely based on a description of the target classes. 
Early works demonstrated how to train zero-shot classifiers based on attributes~\cite{visual_attribute} or numerical descriptors~\cite{zero_data_learning}.
Another approach, which we adopt in this work, is to learn an alignment between image and text embedding spaces~\cite{devise,vse,karpathy,vse-pooling,virtex,convirt}. 
This approach has demonstrated that with modern architectures, contrastive learning, and large data sources it is possible to obtain performance that is competitive with the classical two-step approach that involves fine-tuning on the downstream data~\cite{clip,align}. 
Other efforts in this direction explore image-text alignment or masked language (or image region) modeling~\cite{VisualBERT,uniter}. 
The models have been applied to diverse downstream tasks, including visual question answering~\cite{vqa2}, visual commonsense reasoning~\cite{vcr} and image captioning~\cite{ViLBERT,VL-BERT,12-in-1}. 

\textit{Contrastive learning} techniques are another closely-related research direction.
The high-level idea of a contrastive loss is to simplify the learning task by requiring the model to select the correct answers out of a finite set of carefully designed options.
Intuitively, this simplification of the task may encourage the model to focus on high-level information in an image instead of generic information, resulting in high quality learned representations.
Early works that investigate very specific instances of this idea include~\cite{rel-patch-location,jigsaw}. More recently, contrastive learning was formulated and studied in more general settings~\cite{cpc,simclr,moco}, leading to very promising results. Finally, \cite{clip,align} use contrastive learning for learning from image-text data and derive state-of-the-art zero-shot image classifiers.

% ===============================================
% ===============================================

\section{Methods}\label{sec:methods}

% ===============================================
% ===============================================

\subsection{Contrastive pre-training}\label{sec:contrastive_pre_train}

Collections of images (potentially noisily) paired with free-form text descriptions have emerged as a powerful resource for training visual models. The key advantage therein is that it is not limited by a finite set of predefined categories and instead describes images using open-ended natural language. As a result, models learned from this data can serve as zero-shot learners for a wide range of tasks, \eg classification and image/text retrieval.

Contrastive pre-training is one particularly effective approach for training models from image-text data, which was recently proven to work well in practice~\cite{clip,align}. We take a closer look at this approach and propose a simple, yet highly effective recipe to significantly enhance contrastive pre-training from image-text data.

The key idea behind the contrastive pre-training approach is to learn two embedding models: an image model and a text model, both of which produce representations of the same dimensionality. These models are trained using a contrastive loss. This loss encourages corresponding image-text pairs to have similar embeddings and, conversely, encourages non-corresponding pairs to have distinct embeddings. See~\cite{clip,convirt} for the detailed discussion of the contrastive loss function.

An important detail of this loss function is whether the loss is computed on each accelerator device independently and then accumulated or computed jointly across all devices. 
We ablate this design choice (Appendix~\ref{appendix:batch_size}) and confirm that the latter~\cite{clip,align} consistently results in better performance. We therefore use the global loss in all our experiments and ablations. 

After image and text towers are trained, they can be readily used for zero-shot classification: class names or descriptions are embedded with the text model. Then, for a given image the label is selected that has the embedding closest to the embedding of the image. This approach also works for image-text retrieval.

% ===============================================
% ===============================================

\subsection{Contrastive-tuning}\label{sec:contrastive_tuning}

\begin{figure}[t]
    \centering
    \includegraphics[width=\linewidth]{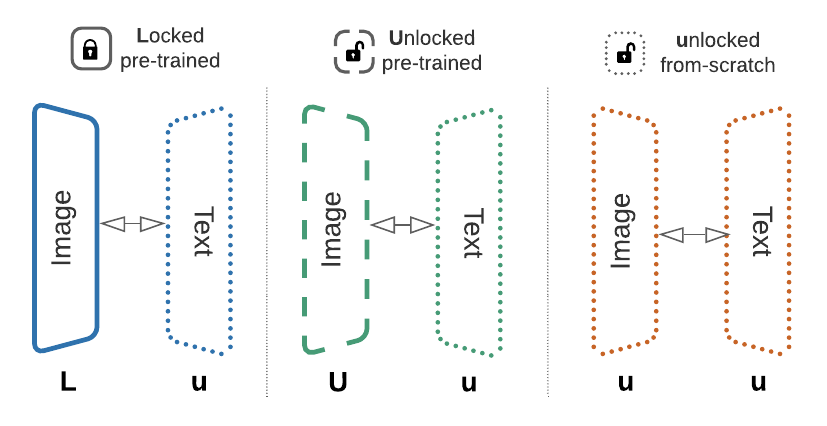}
    \vskip-1em
    \caption{Design choices for contrastive-tuning on image-text data. Two letters are introduced to represent the image tower and text tower setups. {\tt L} stands for locked variables and initialized from a pre-trained model, {\tt U} stands for unlocked and initialized from a pre-trained model, {\tt u} stands for unlocked and randomly initialized. 
    \Lu is named as “Locked-image Tuning” (\lit{}).
    }
    \label{fig:teaser_design}
\end{figure}

Contrastive pre-training can be viewed as learning two tasks at the same time: (1)~learning an image embedding and (2)~learning a text embedding to align with the image embedding space. While contrastive pre-training on image-text data works well for solving both of these tasks simultaneously, it may be not the optimal approach.

When not using contrastive pre-training on image-text data, a standard approach to learning image embeddings is to use a large and relatively clean dataset of (semi)-manually labeled images. Large scale and and high quality of such data result in state-of-the-art image embeddings. Some dataset choices for learning powerful image embeddings are ImageNet-21k~\cite{imagenet}, JFT-300M~\cite{unreasonable_effectiveness_of_data}. 

However, this common approach has a clear weakness: it is limited to a \textit{predefined set of categories} and, thus, the resulting models can only reason about these categories. In contrast, image-text data does not have this limitation, as it learns from the \textit{free-form text} that potentially spans a broad range of real-life concepts. On the other hand, image-text data that is available may be of lower quality (for learning image embeddings) than carefully curated datasets.

We propose \textit{contrastive-tuning} to combine advantages of both sources of data. One specific way of doing this is to initialise the contrastive pre-training with an image model that was \textit{already pre-trained} using cleaner \mbox{(semi-)}manually labeled data. This way the image-text alignment is learned independently of image embedding, enabling benefit from both data sources. 

Beyond using supervised pre-trained image models, the proposed contrastive-tuning is also flexible enough to integrate any models that can produce meaningful representations. We verify this in our experiments using self-supervised pre-trained image models.

Similar lines of reasoning can also be applied to the text tower, as there are many powerful pretrained models that use text-specific data sources and learning techniques. 

% ===============================================
% ===============================================

\subsection{Design choices and Locked-image Tuning}
\label{sec:contrastive_design}

Introducing pre-trained image or text models into the contrastive learning setting involves several design choices. First, each tower (image and text) can independently be initialized randomly or from a pre-trained model. For a pre-trained model there are at least two variants: we can lock (freeze) it or allow fine-tuning. Note that there are many choices between these two extremes (\eg partial freezing of selected layers, or custom learning rates), but
they are not investigated in this paper.  

Pre-trained image-text models may have different representation sizes, while the contrastive loss expects representations of the same size. 
To compensate, we add an optional linear projection (head) to each tower, which maps the representations to a common dimensionality. 
Preliminary investigations with tried MLP-based heads did not yield significant improvements over such a simple linear head.

We introduce a two-character notation to discuss the potential design choices outlined above (see Figure~\ref{fig:teaser_design}).
Each character encodes the setting chosen for the image model and the text model (in this order).
We define three potential settings: `L` (locked weights, a initialized from pre-trained model), `U` (unlocked/trainable weights, initialized from a pre-trained model) and `u` (unlocked/trainable weights, randomly initialized).
For example, the notation \Lu means locked pre-trained image model, and unlocked (trainable) randomly initialized text model. Previous works training models from scratch~\cite{clip,align} are \uu. In our experiments we find the \Lu setting to work particularly well, so we explicitly name it as \textit{Locked-image Tuning} (\litf{}).

% ===============================================
% ===============================================

\section{Image-text datasets}\label{sec:data}

\textbf{CC12M.} The Conceptual Captions dataset~\cite{cc3m} extracts, filters 
\& transforms image \& alt-text pairs from web pages.
We use the latest 12 million image-text pair version, \ie CC12M~\cite{cc12m}.
Due to expired URLs, only 10 million image-text pairs were used for our experiments.

\textbf{YFCC100m.} The Yahoo Flickr Creative Commons dataset~\cite{yfcc100m} contains 100 million media objects. 
Of these, 99.2 million are photos that come with rich metadata including camera info, timestamp, title, description, tags, geolocation, and more. \cite{clip} defines and uses a subset of 15 million images that have been filtered for English text of high quality, which we call YFCC100m-CLIP. A detailed investigation of this dataset and how best to use it, including whether to filter it, is presented in Appendix~\ref{appendix:yfcc}.

\textbf{Our dataset.} We collect 4 billion image and alt-text pairs following the same process as ALIGN~\cite{align}, with the same image-based filtering but simpler text-based filtering.
Appendix~\ref{appendix:align_vs_our} shows that reducing text filtering does not harm performance.
To avoid misleading evaluation results, we remove from our dataset near-duplicate images of all splits from all datasets we evaluate on. 
We do not consider the creation of our dataset a main contribution of this paper; we just simplify the data collection process in ALIGN~\cite{align} to demonstrate the efficacy of our methods at scale.

% ===============================================
% ===============================================

\section{Experiments}\label{sec:exps}

In this section, we first compare \litf to state-of-the-art image-text models.
We consider two scenarios: (1)~only using public datasets for model training and (2)~using privately gathered data.
We then present 
learnings from 
experimental evaluations of contrastive tuning design choices with various training settings \& datasets. 
We generally perform evaluation on 0-shot ImageNet classification (“0-shot”) and MSCOCO image (“T$\rightarrow$I”) and text (“I$\rightarrow$T”) retrieval.

% ===============================================
% ===============================================

\subsection{Comparison to the previous state-of-the-art}
\label{sec:sota}

\begin{table}[t]
  \newcolumntype{C}{>{\centering\arraybackslash}X}
  \newcolumntype{R}{>{\raggedleft\arraybackslash}X}
  \setlength{\tabcolsep}{0pt}
  \setlength{\extrarowheight}{5pt}
  \renewcommand{\arraystretch}{0.75}
  \centering
  % Format is type{width}, p: paragraph, top-align, C is custom, c is squeezed.
  \begin{tabularx}{\linewidth}{p{0.1cm}p{0.5cm}Cp{1.5cm}Cp{0.1cm}Cp{0.1cm}Cp{0.1cm}Cp{0.1cm}Cp{0.1cm}Cp{0.1cm}C}
    \toprule[1pt]
\bf{\multirow{3}{*}{\rotatebox{90}{\hspace*{-2pt}Dataset}}}  &&  && \multirow{3}{*}{\rotatebox{90}{\hspace*{0pt}INet}} && \multirow{3}{*}{\rotatebox{90}{\hspace*{0pt}INet-v2}} && \multirow{3}{*}{\rotatebox{90}{\hspace*{0pt}INet-R}} &&
    \multirow{3}{*}{\rotatebox{90}{\hspace*{0pt}INet-A}} &&
    \multirow{3}{*}{\rotatebox{90}{\hspace*{0pt}ObjNet}} && \multirow{3}{*}{\rotatebox{90}{\hspace*{0pt}ReaL}} &&
    \multirow{3}{*}{\rotatebox{90}{\hspace*{0pt}VTAB-N}} \\
    && \bf{Method} &&\\
    \\
    \midrule
    \multirow{3}{*}{\rotatebox{90}{\hspace*{-2pt}Private}} && CLIP~\cite{clip} && 76.2 && 70.1 && 88.9 && 77.2 && 72.3 && - && - \\
     && ALIGN~\cite{align} && 76.4 && 70.1 && 92.2 && 75.8 && - && - && -\\
     && \mbox{\textit{\lit{}}} && \bf{85.2} && \bf{79.8} && \bf{94.9} && \bf{81.8} && \bf{\objsota{}} && 88.6 && 74.7\\
     \midrule
     \multirow{3}{*}{\rotatebox{90}{\hspace*{-2pt}Public}} && CLIP~\cite{clip} && 31.3 && - && - && - && - && - && -\\
     && OpenCLIP~\cite{openclip} && 34.8 && 30.0 && - && - && - && - && - \\
     % go/lit-oss-fb
     && \mbox{\textit{LiT}} && \bf{75.7} && \bf{66.6} && 60.4 && 37.8 && 54.5 && 82.1 && 63.1\\
     \midrule
     \multirow{2}{*}{\rotatebox{90}{\hspace*{9pt}*}} && ResNet50~\cite{resnet} && 75.8 && 63.8 && 36.1 && 0.5 && 26.5 && 82.5 && 72.6 \\
    \bottomrule[1pt]
  \end{tabularx}
  \caption{Zero-shot transfer accuracies (\%) on ImageNet,  five OOD test variants, and seven VTAB-natural tasks. Results are reported on both public datasets and privately gathered data. For reference, we include the ResNet50 model pre-trained on ImageNet, supervised fine-tuned on downstream datasets. We use * to denote multiple datasets during supervised fine-tuning.
  }\label{table:sota}
\end{table}

In this section, we present \lit{} results on our dataset.
The image tower is initialized with a ViT-g/14 model\footnote{An earlier version of this paper reported slightly lower numbers with the ViT-g/14 model, e.g. ImageNet accuracy was 84.5\% vs 85.2\%. We fixed a model loading bug with ViT-g/14 in this version. Other results are not affected.} pre-trained on JFT-3B\cite{vitg}, which has been de-duplicated against the downstream tasks. 
We use $32$k batch size, and tune for $18$ billion image-text pairs seen (roughly $550$k steps). See Appendix~\ref{appendix:our_details} for details.

We compare the LiT method with the previous state-of-the-art methods, including CLIP~\cite{clip} and ALIGN~\cite{align}. 
In Table~\ref{table:sota}, we report zero-shot classification results on the ImageNet dataset, five out-of-distribution test variants and seven VTAB-natural tasks~\cite{vtab}.
Our model significantly outperforms the previous state-of-the-art methods at ImageNet zero-shot classification.
The 9\% and  8.8\% improvement over CLIP and ALIGN, respectively, halves the gap between zero-shot transfer results and supervised fine-tuned results~\cite{vitg,coatnet}.

\textbf{Robustness.}
We evaluate robustness on ImageNet-v2~\cite{imagenet_v2}, -R~\cite{imagenet_r,imagenet_sketch}, -A~\cite{imagenet_a}, -ReaL~\cite{imagenet_real}, and ObjectNet~\cite{objectnet}, following CLIP and ALIGN. 
On all of the OOD variants, our model consistently outperforms the previous models. 
Notably, the \lit{} model sets a new state-of-the-art \objsota{}\% accuracy on the ObjectNet test set.
The pre-trained ViT-g/14 model~\cite{vitg}, achieves 70.5\% accuracy on the ObjectNet test set when fine-tuned on ImageNet.
This model gets more than 10\% improvement when instead locked-image tuned (LiT) on our image-text dataset.

\textbf{Diverse downstream tasks.}
We evaluate the LiT models on VTAB, consisting of 19 diverse tasks.
We report averaged results on seven VTAB-natural tasks in Table~\ref{table:sota}. The LiT models achieve promising zero-shot results, comparing to the supervised fine-tuned ResNet50 baseline. 
In Appendix~\ref{appendix:vtab}, we present zero-shot transfer details on VTAB, as well as more results and analysis on the specialized tasks and structured tasks.

\textbf{Data \& compute efficiency.} Figure~\ref{fig:teaser_curve} shows more results when tuning with fewer seen image-text pairs. 
With \lit{} the model achieves 81.7\% top-1 accuracy on 0-shot ImageNet transfer, with only 300M image-text pairs seen.
In comparison, it took the from-scratch method (\ie CLIP) 12.8B image-text pairs seen, \ie 40 times more data pairs, to reach 76.2\% top-1 accuracy. 
With a pre-trained image model, the proposed setup converges significantly faster than the standard from-scratch setups reported in the literature.
\lit{} provides a way to reuse the already pre-trained models in the literature, amortizing the computational resources used to re-generate the image models.

\textbf{Results on public datasets.}
Given high data efficiency of \lit, we investigate how well it performs when using only smaller, publicly available models and datasets.
Specifically, we tune an ImageNet-21k pre-trained ViT-L/16 model~\cite{augreg} on the union of the
\emph{YFCC100m-CLIP} and \emph{CC12M} datasets. More details of the training setup are provided in Appendix~\ref{appendix:public_details}.
As a result we achieve unprecedented \textbf{75.7\%} zero-shot transfer on ImageNet, an absolute improvement of 30.9\% over the previously reported state-of-the-art result~\cite{openclip} that uses only public data sources. We also obtain strong results on a wide range of robustness datasets and the VTAB-natural tasks, see Table~\ref{table:sota}. 

% ===============================================
% ===============================================

\subsection{Evaluation of design choices}
\label{sec:design_choices}

\begin{table}[t]
  \newcolumntype{C}{>{\centering\arraybackslash}X}
  \newcolumntype{R}{>{\raggedleft\arraybackslash}X}
  \setlength{\tabcolsep}{0pt}
  \setlength{\extrarowheight}{5pt}
  \renewcommand{\arraystretch}{0.75}
  \centering
  % Format is type{width}, p: paragraph, top-align, C is custom, c is squeezed.
  \begin{tabularx}{\linewidth}{p{1.5cm}p{0.1cm}Cp{0.1cm}Cp{0.1cm}Cp{0.1cm}C}
    \toprule[1pt]
     \bf{Method} && \bf{ImgNet} && \bf{ImgNet-v2} && \bf{Cifar100} && \bf{Pets}\\
    \midrule
     \Lu && 70.1 && 61.7 && 70.9 && 88.1 \\
     \Uu && 57.2 && 50.2 && 62.1 && 74.8 \\
     \uu && 50.6 && 43.3 && 47.9 && 70.3 \\
    \bottomrule[1pt]
  \end{tabularx}
  \caption{Evaluation of design choices on our large dataset.}\label{table:design_our}
\end{table}

\begin{figure}[t]
    \centering
    \includegraphics[width=\linewidth]{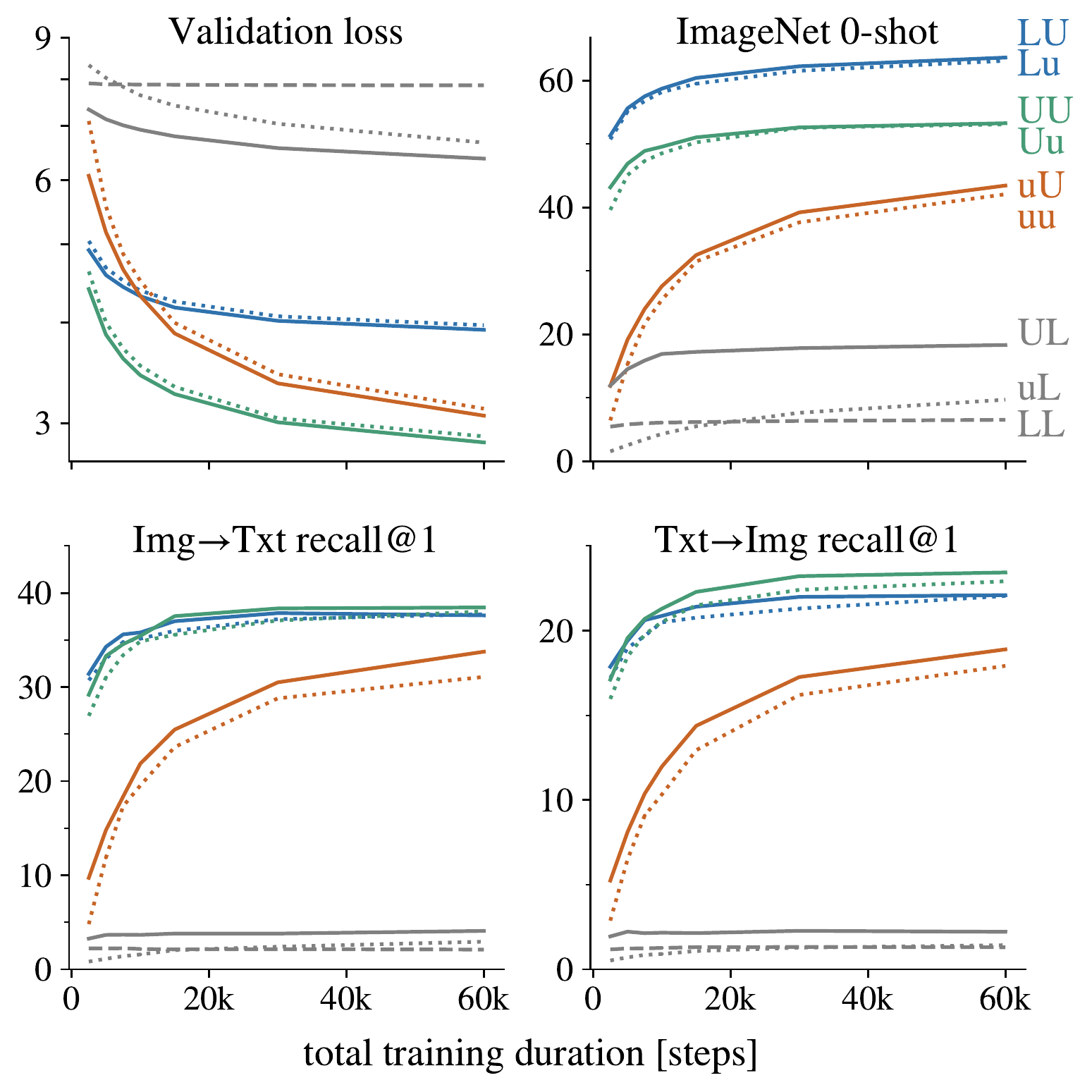}
    \caption{An in-depth study of the possible locking and initialization settings of \protect\lit on the YFCC100m-CLIP dataset. A pre-trained image tower works best, while pre-training of the text tower only helps a little. These are \textbf{not} training curves; each point is the final value reached by a training run of that duration.}
    \label{fig:design}
\end{figure}

\textbf{Small-scale thorough investigation.} We first perform an in-depth study on various combinations of the image and text towers being initialized with pre-trained weights and locked (`L`) or unlocked (`U`) or being randomly initialized and unlocked (`u`).
We train each setting many times on the YFCC100m-CLIP dataset, varying the total number of steps from $2\,500$ to $60\,000$ in order to understand the setting's trajectory, and sweeping over learning-rates and weight-decays to avoid being misled.
Details can be found in Appendix~\ref{appendix:public_details}.
Figure~\ref{fig:design} shows the best result for each setting for each duration, \ie each point on the curves is a separate full run for that duration.
It is evident that locking the image tower almost always works best and using a pre-trained image tower significantly helps across the board, whereas using a pre-trained text tower only marginally improves performance, and locking the text tower does not work well.

\textbf{This still holds in the near-infinite data regime.} One may hypothesize that locking the pre-trained image tower only helps because the YFCC100m-CLIP dataset is \emph{relatively} small (15 million images, compared to 400M~\cite{clip} or 1.8B~\cite{align}), and that a randomly initialized image tower will eventually outperform a locked one on much larger image-text datasets.
The trajectory of the \Uu and \UU settings in Figure~\ref{fig:design} may seem to support this expectation.

Maybe surprisingly, 
%in this section we 
experimental results
show that this is not the case, and locking the image tower provides benefits even when contrastively tuning on a very large dataset of image-text pairs.
Table~\ref{table:design_our} shows results of contrastive tuning on our dataset of 4 billion images in three settings: \Lu, \Uu, and \uu.
Implementation details can be found in Appendix~\ref{appendix:our_details}.
The from-scratch method \uu unsurprisingly achieves better performance than with smaller datasets such as CC12M and YFCC100m-CLIP. 

Initializing the image tower from a pre-trained model provides even better performance and is a relatively straightforward extension of CLIP/ALIGN.
Perhaps surprisingly, the frozen setup \Lu, achieves even better results. 
While potentially counter-intuitive, another perspective is that \lit simply learns a text tower that extracts knowledge from a strong image embedder. 
This flexible \& performant setup can turn existing vision backbones into a zero-shot learners, by attaching a text-embedding tower.

\begin{figure}[t]
    \centering
    \includegraphics[width=\linewidth]{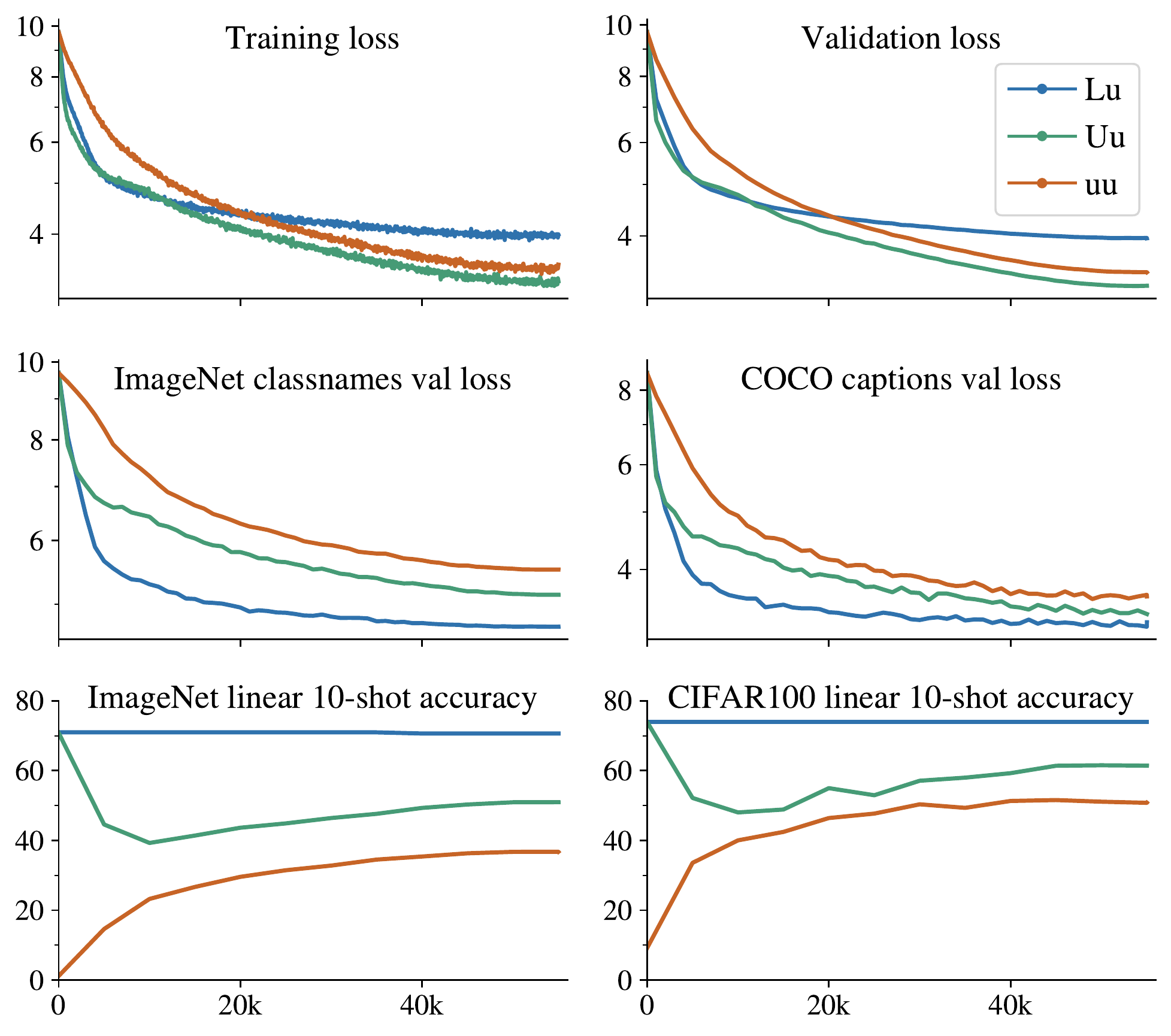}
    \caption{Comparing the loss on the dataset used for \protect\lit (top row) to the loss on out-of-distribution (zero-shot) datasets (middle row) and the “representation quality” as measured by linear few-shot evaluation on the pre-logits (bottom row).
    This reveals how the different settings behave, see text for details.%further interpretation.
    }
    \label{fig:design_loss_repr}
\end{figure}

\textbf{Why is locked ({\tt \textbf{L}}) better than unlocked ({\tt \textbf{U}})?} It is somewhat surprising and counter-intuitive that locking the image tower works better than allowing it to adapt during the contrastive-tuning; Figure~\ref{fig:design_loss_repr} gives hints as to why.

The first row shows that locking the image tower leads to substantially worse (contrastive) loss on the dataset used for \lit, while the loss of the locked image variant is substantially better on out-of-distribution datasets such as COCO captions (middle row).

We also measure the \emph{representation quality} of the image model (bottom row) via the performance achieved by a few-shot linear regression on its pre-logits, as is commonly done in the self-supervised representation learning literature.
Taken together, these figures reveal that the image representation of a pre-trained image model generalizes very well, but contrastively fine-tuning it worsens the generality of the visual representation, leading it to be better on the contrastive dataset, but worse everywhere else.
This indicates that locking the image tower during tuning, \ie \lit, \emph{leads to a text model that is well aligned to an already strong and general image representation}, as opposed to an image-text model that is well aligned but specialized to the dataset used for alignment.

Intermediate variants, such as first locking and later unlocking the image tower or separating learning-rates are explored in Appendix~\ref{appendix:lr_schedules}; we did not find a strictly better setup than \lit and leave this as an open research question.

% ===============================================
% ===============================================

\subsection{\protect\lit works better for more generally pre-trained models}
\label{sec:more_image_pretrainings}

One may believe that \lit only works because the image tower is initialized with a backbone that was supervisedly pre-trained for classification, and hence remains a supervised classifier, as opposed to becoming an image-text model.
We design a controlled experiment to verify whether that is the case. We find that on the contrary, more generally pre-trained models are better suited for \lit.

We select a set of image models that all use the same ViT-B/16 architecture but were pre-trained in various ways: supervised (AugReg~\cite{augreg}) on ImageNet (IN), on the large but narrow Places~\cite{places365} dataset, on the much broader ImageNet-21k (IN21k), or fully unsupervised (DINO and MoCo-v3).
All but the Places model achieve similar ImageNet top-1 accuracies of around 77\% as reported in their respective publications, and can thus be considered \emph{similarly good} models.

\begin{table}[t]
  % \small
  \newcommand{\hi}{\cellcolor[rgb]{0.95,1.0,0.95}}
  \newcommand{\mi}{\cellcolor[rgb]{1.0,1.0,0.95}}
  \newcommand{\lo}{\cellcolor[rgb]{1.0,0.95,0.95}}
  \newcolumntype{C}{>{\centering\arraybackslash}X}
  \newcolumntype{R}{>{\raggedleft\arraybackslash}X}
  \setlength{\tabcolsep}{0pt}
  \setlength{\extrarowheight}{5pt}
  \renewcommand{\arraystretch}{0.75}
  \centering
  % Format is type{width}, p: paragraph, top-align, C is custom, c is squeezed.
  \begin{tabularx}{\linewidth}{p{2cm}p{0.1cm}Cp{0.1cm}Cp{0.1cm}Cp{0.1cm}Cp{0.1cm}Cp{0.1cm}Cp{0.1cm}C}
    \toprule[1pt]
     \multirow{3}{=}[-20pt]{\shortstack[l]{\bf{Model:}\\ ViT-B/16}} &&
     \multicolumn{7}{c}{\bf{Pre-training}} &&
     \multicolumn{5}{c}{\bf{LiT}} \\
     \cmidrule[0.5pt]{3-9} \cmidrule[0.5pt]{11-15}
     &&
     {\centering \rotatebox{90}{Dataset}} &&
     {\centering \rotatebox{90}{Labels?}} &&
     {\centering \rotatebox{90}{Full IN}} &&
     {\centering \rotatebox{90}{10-shot}} &&
     {\centering \rotatebox{90}{0-shot}} &&
     {\centering \rotatebox{90}{I$\rightarrow$T}} &&
     {\centering \rotatebox{90}{T$\rightarrow$I}}\\
    \midrule
     \mbox{MoCo-v3~\cite{mocov3}} && IN    && n && \hi76.7 && \hi60.6 && \hi55.4 && \hi33.5 && \hi17.6 \\
     DINO~\cite{dino}      && IN    && n && \hi78.2 && \hi61.2 && \hi55.5 && \hi33.4 && \hi18.2 \\
     AugReg~\cite{augreg}  && IN21k   && y && \hi77.4 && \hi63.9 && \hi55.9 && \hi30.3 && \hi17.2 \\
    \arrayrulecolor{lightgray}\midrule[0.25pt]\arrayrulecolor{black}
     AugReg~\cite{augreg}  && IN    && y && \hi77.7 && \lo77.1 && \lo64.3 && \lo25.4 && \lo13.8 \\
    % \arrayrulecolor{lightgray}\midrule[0.25pt]\arrayrulecolor{black}
     AugReg~\cite{augreg}  && Places && y && \lo- && \lo22.5 && \lo28.5 && \lo25.1 && \lo12.9 \\
    \bottomrule[1pt]
  \end{tabularx}
  \caption{The role of pre-training method for the image model: as long as it is general, it does not matter. The background coloring denotes whether a value is \colorbox[rgb]{0.95,1.0,0.95}{similar} or \colorbox[rgb]{1.0,0.95,0.95}{far away} from the others in that column.}\label{table:more_image_pretrainings}
\end{table}

Table~\ref{table:more_image_pretrainings} shows model performance without \lit (ImageNet 10-shot, and accuracy when fully fine-tuned on ImageNet) alongside achieved performance with \lit on YFCC100m-CLIP (zero-shot ImageNet classification and MS Coco retrieval).

From these results, we conclude that models which are pre-trained in a generic way (\eg on large amounts of data, or in an unsupervised way) and have similar representation quality, become similarly good image-text models after locked-image tuning (\lit).
However, this also shows that a narrowly pre-trained model (AugReg-IN and AugReg-Places) will perform misleadingly well on its narrow task (0-shot IN for AugReg-IN), but significantly fall behind on more general image-text tasks (MSCOCO captions).
These findings highlight the importance of a generally pre-trained model and varied set of evaluation tasks.

\textbf{Is this specific to ViT image models?} No. Here we fixed the architecture to avoid confounders, but Appendix~\ref{sec:more_image_architectures} explores other architectures.

% ===============================================
% ===============================================

\subsection{Which text model to use?}
\label{sec:more_text_models}

\begin{table}[t]
  \newcolumntype{C}{>{\centering\arraybackslash}X}
  \newcolumntype{R}{>{\raggedleft\arraybackslash}X}
  \setlength{\tabcolsep}{0pt}
  \setlength{\extrarowheight}{5pt}
  \renewcommand{\arraystretch}{0.75}
  \centering
  % Format is type{width}, p: paragraph, top-align, C is custom, c is squeezed.
  \begin{tabularx}{\linewidth}{p{0.4cm}p{0.1cm}p{1.2cm}p{0.1cm}p{0.5cm}p{0.1cm}Cp{0.1cm}Cp{0.1cm}C}
    \toprule[1pt]
     && \textbf{Model} && \textbf{Tok} && \textbf{INet 0shot} && \textbf{I$\rightarrow$T} && \textbf{T$\rightarrow$I} \\
    \midrule
    \multirow{5}{*}{\rotatebox{90}{YFCC-CLIP}} && ViT  && SP && 57.2 \phantom{(+0.0)} && 29.7 \phantom{(+0.0)} && 16.9 \phantom{(+0.0)} \\
     && T5   && SP && 57.8 (+1.4)           && 29.4 (+1.6)           && 17.2 (+1.2) \\
     && mT5  && SP && 58.1 (+1.2)           && 28.3 (+0.4)           && 16.4 (+1.0) \\
     && BERT && WP && \textbf{58.8} (+0.7)  && \textbf{35.2} (+1.1)  && \textbf{20.0} (+0.7) \\
     && ViT  && WP && 56.4 \phantom{(+0.0)} && 28.2 \phantom{(+0.0)} && 17.3 \phantom{(+0.0)} \\
    \arrayrulecolor{lightgray}\midrule[0.25pt]\arrayrulecolor{black}
     \multirow{3}{*}{\rotatebox{90}{Ours}} && ViT  && SP && 68.8 \phantom{(+0.0)} && 43.6 \phantom{(+0.0)} && 28.5 \phantom{(+0.0)} \\
     && ViT  && WP && 68.8 \phantom{(+0.0)} && 45.4 \phantom{(+0.0)} && 29.7 \phantom{(+0.0)} \\
     && BERT && WP && 65.8 \phantom{(+0.0)} && 43.8 \phantom{(+0.0)} && 28.6 \phantom{(+0.0)} \\
    \bottomrule[1pt]
  \end{tabularx}
\caption{The effect of different text encoders on zero-shot performance. The  main  numbers  show  performance  achieved  when the text tower is randomly initialised; the numbers in brackets are the further improvement achieved when the text tower is initialized with a pre-trained language model.
% The additional improvement when using pre-trained weights is given in parentheses.
The \emph{Tok} column indicates whether a SentencePiece or WordPiece tokenizer was used.}\label{tab:text_encoder}
\end{table}

While related work has so far focused on the image model, the text model plays an important yet 
underexplored role in contrastive image-text learning. We consider four possible transformer-based text models~\cite{transformer}---the transformer from ViT-B~\cite{vit} which also resembles that used in CLIP~\cite{clip}, T5-base~\cite{T5}, mT5-base~\cite{mt5}, and the classic BERT-base~\cite{bert}---and whether to initialise them randomly, or from a pre-trained checkpoint. BERT uses a WordPiece (WP) tokenizer~\cite{translate,japkorvoice}, and all others use the SentencePiece (SP) tokenizer~\cite{sentencepiece}, a component which we also ablate with the ViT model.

Table~\ref{tab:text_encoder} shows the results of \lit{} using an AugReg-ViT-B/32 on YFCC100M-CLIP and our dataset using the \emph{base} sized variant of these text models.
We sweep over various learning-rates and weight-decays separately for each combination to avoid being misled.
Our observations differ slightly between the \emph{relatively} small YFCC100m-CLIP dataset, and our much larger dataset, we first discuss the former.
First, we see a small but consistent improvement by initializing the text model with pre-trained weights.
Second and somewhat unexpectedly, we find that the BERT model performs significantly better than others, especially for retrieval.
In order to disentangle the contribution of the architecture from the tokenizer, we further apply \lit using a ViT text encoder paired with BERT's WordPiece tokenizer and see no improvement.
We believe that small differences in the architecture, such as initialization and LayerNorm placement, are responsible for the slightly better generalization of BERT that we observe. However, we also found the BERT model to be less stable to train.
For the large-scale experiments on our dataset, we do not observe this improvement anymore, and favor sticking with the more stable ViT SentencePiece combination.

\textbf{What about model capacity?} Previous works used relatively low-capacity text models. We show in Appendix~\ref{sec:model_capacity_impact} that increasing the text tower's capacity consistently improves performance. The same is true, and more pronounced, for the image tower.

% ===============================================
% ===============================================

\subsection{Do duplicate examples matter for \protect\lit?}
\label{sec:deduplication}

\begin{table}[tb]
  \newcolumntype{C}{>{\centering\arraybackslash}X}
  \newcolumntype{R}{>{\raggedleft\arraybackslash}X}
  \setlength{\tabcolsep}{0pt}
  \setlength{\extrarowheight}{5pt}
  \renewcommand{\arraystretch}{0.75}
  \centering
  % Format is type{width}, p: paragraph, top-align, C is custom, c is squeezed.
  \begin{tabularx}{\linewidth}{p{1.5cm}p{0.1cm}Cp{0.1cm}Cp{0.1cm}Cp{0.1cm}Cp{0.1cm}C}
    \toprule[1pt]
     \bf{Dedup} && \bf{\#tune} && \bf{\#eval} && \bf{ImgNet} && \bf{I$\rightarrow$T} && \bf{T$\rightarrow$I}\\
    \midrule
     - && 0 && 0 && 70.2 && 43.6 && 28.4 \\
     test && 2.6M && 76K && 70.2 && 43.3 && 28.3 \\  % 2_575_200
     train+test && 3.6M && 220K && 69.9 && 43.7 && 28.4 \\  % 3_637_630
    \bottomrule[1pt]
  \end{tabularx}
  \caption{Results on various de-duplication setups. \#tune images are removed from the \lit dataset due to \#eval images in the evaluation datasets. We report results averaged across three runs.}\label{table:duplicates}
\end{table}

One relevant question in the context of large-scale training is the role of duplicate examples between upstream datasets and downstream datasets. 
We answer this question by performing experiments on three different upstream de-duplication setups: (1)~no de-duplication; (2)~de-duplicate against downstream test splits only; (3)~de-duplicate against downstream train and test splits.
We conduct experiments using the \Lu setup on our dataset.
We use a B/32 image model pre-trained on the JFT-3B dataset~\cite{vitg}, which has been de-duplicated against downstream train and test splits.

In Table~\ref{table:duplicates}, we show the number of duplicate samples found between upstream datasets and downstream datasets during de-duplication. 
In the de-duplication process, a downstream image may have multiple upstream duplicate examples, \eg due to image copies on the web. 
As a result, the number of duplicate examples on the upstream dataset is significantly larger than the number on the downstream datasets.
The downstream number indicates how many downstream images had a duplicate detected, while the upstream number indicates how many images are removed from the image-text dataset.

We apply \lit{} on the three setups, and the zero-shot transfer results vary little.
More results with larger backbone can be found in Appendix~\ref{appendix:more_dedup}, with consistent conclusions.
It indicates that the duplication of examples here \textit{does not} influence the results strongly.
This observation is also consistent with previous conclusions~\cite{bit,clip}.
A possible interpretation is that with a large upstream dataset, the model may not memorize those duplicate examples. 

Throughout this paper, we report results using the strictest setup (3) with proper de-duplication against downstream train splits and test splits, to avoid data leakage.

% ===============================================
% ===============================================

\subsection{Technical advantages of locked image models}
\label{sec:precompute}

Besides potential modelling advantages previously explored, using a locked image tower has several more benefits.
First, the training is significantly sped-up and memory use reduced as no gradients are computed for the image tower.
Second, if no augmentations are used, such as in our large-data experiment, the image model's embeddings can be precomputed once, further reducing computation time and memory requirements.
Appendix~\ref{appendix:precompute} shows concrete measurements.
Taken together, these implementation features unlock the use of enormous models at very large batch-sizes.

% ===============================================
% ===============================================

\subsection{Preliminary multilingual experiments}
\label{sec:multilingual}

It is currently common practice~\cite{align,clip} to filter image-text datasets to English language data only%
.
We believe that removing this restriction has the potential to benefit a larger part of the world's population.
Concurrent work~\cite{mural} has relied on additional translated text pairs for training the text encoder.
In contrast, we do not require any translations and purely rely on the pre-trained, locked image model to bridge the language barrier.
In this section, we report preliminary experiments that show the promise of \lit for multilingual image-text models.

\begin{figure}[t]
    \centering
    \includegraphics[width=0.9\linewidth]{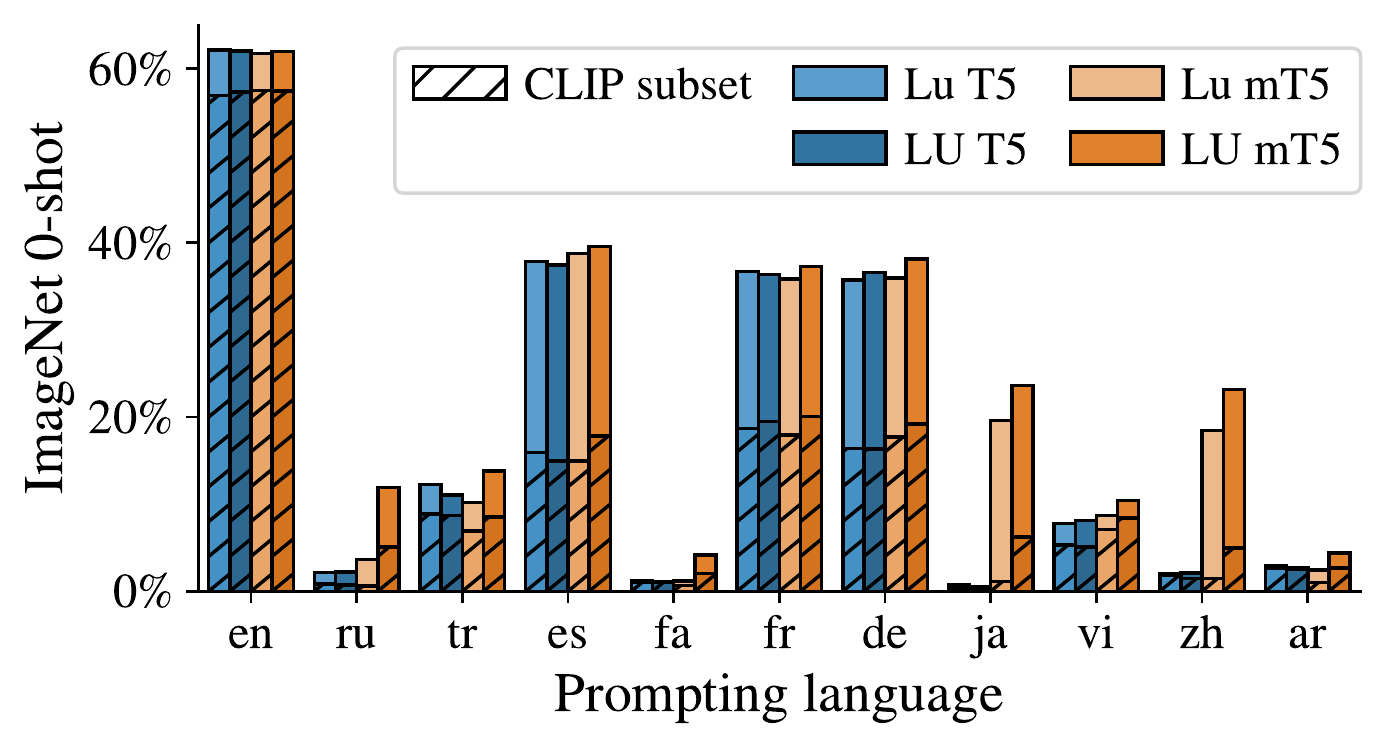}
    \caption{Including non-English data unlocks multilingual zero-shot models without hurting English performance. In such a regime, multilingual text pre-training can be more useful for low-resource languages.}
    \label{fig:xlang_i1k}
\end{figure}

We apply \lit{} on an AugReg-i21k ViT-B/32 with the T5~\cite{T5} and mT5~\cite{mt5} base encoders, both with and without the pre-trained checkpoints. We do this on both the full YFCC100m dataset, and the reduced English-only CLIP subset, and we use all available text as supervision signal (See Appendix~\ref{appendix:yfcc}).
We evaluate the resulting model's multilingualism in two ways, both of which have limitations discussed in Appendix~\ref{appendix:multilingual_details}.
First, we translate the ImageNet prompts into the most common languages using an online translation service and perform zero-shot classification in each of them; this evaluation is shown in Figure~\ref{fig:xlang_i1k}.
Second, we use the Wikipedia based Image Text (WIT) dataset~\cite{wit_dataset} to perform T~$\rightarrow$~I retrieval across more than a hundred languages. Figure~\ref{fig:xlang_wit} gives a summary of this evaluation; a more detailed variant is provided in Appendix~\ref{appendix:multilingual_details}.

The high-level conclusions are consistent across both evaluations: training on the full dataset improves performance on non-English languages much more than on English, using a multilingual tokenizer (as in mT5) significantly helps languages that do not use the Latin script, and starting from a pre-trained multilingual text model can further help.
The combination of all three improvements barely has any effect when evaluated in English, but significantly improves performance on the long tail of languages.
This is a promising result for unlocking multimodal models for low-resource languages.

\begin{figure}[t]
    \centering
    \includegraphics[width=\linewidth]{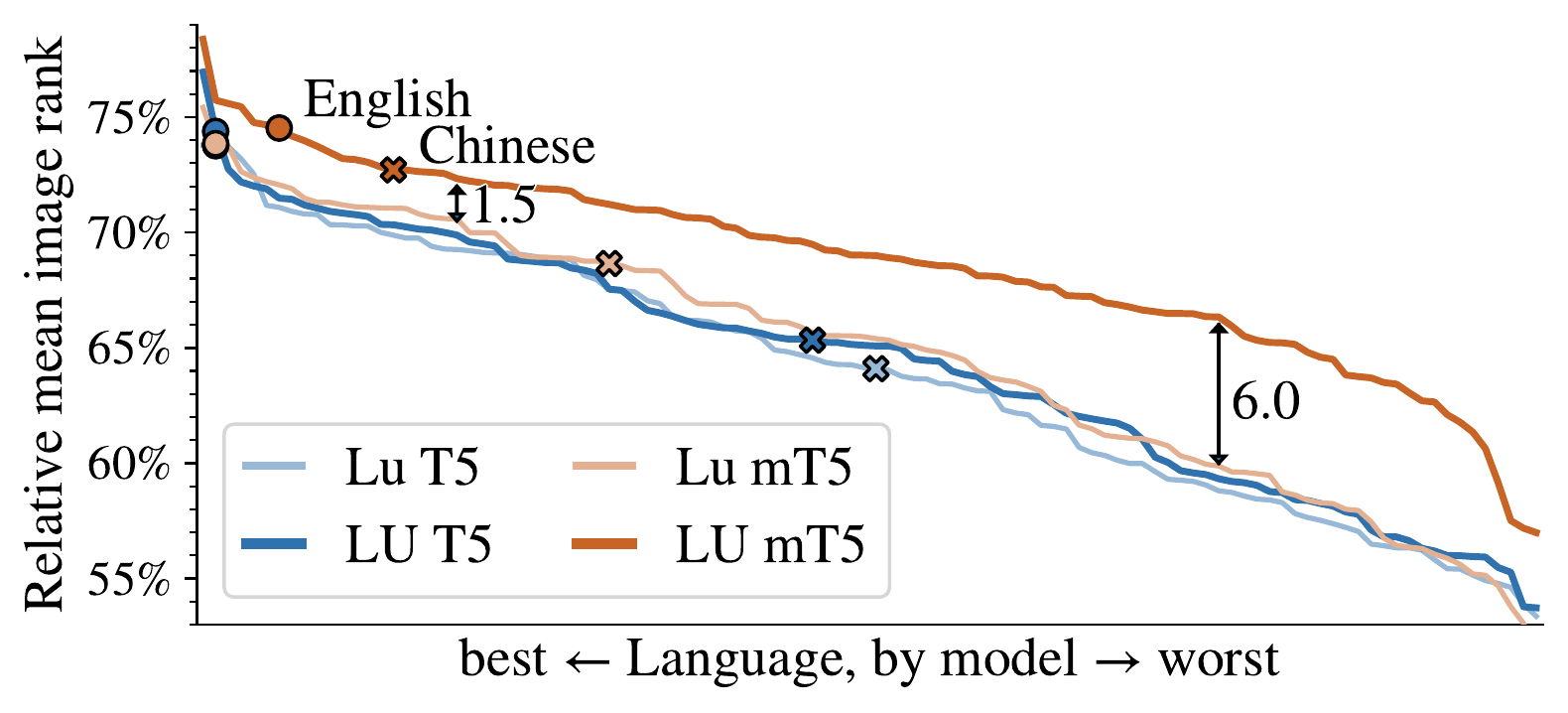}
    \caption{Image retrieval performance over 100 languages reveals that unfiltered data and a multilingually pre-trained text model can significantly increase long-tail performance.}\label{fig:xlang_wit}
\end{figure}

% ===============================================
% ===============================================

\section{Discussion}\label{sec:discussion}

\textbf{Limitations.} This work explores only classification and retrieval as zero-shot transfer tasks.
We leave evaluating zero-shot transfer to a broader set of tasks such as detection, segmentation, visual question answering, and image captioning as future work in order to limit our scope.

On cross-modal retrieval tasks, we have not observed as clear a benefit of the \Lu setup compared to \Uu or \UU (Figure~\ref{fig:design}). 
For very long tuning schedules, \Uu or \UU sometimes overtake \Lu on these tasks.
Our results suggest that the proposed \Lu setup can still save computational cost within a fixed budget, but with a large enough budget, it may be useful to also consider the \Uu setup if zero-shot classification is not the primary end goal.

\textbf{Societal impact.} This work shows how one can easily add a text-tower to a pre-trained image model. While there are many useful applications, like most research, it is a double-edged sword: the technique also makes it simpler to create malicious, offensive, or obscene text tower pendants to existing image models.
Further research is needed on how to best equip open-world image-text models with the behaviour we desire. 

% ===============================================
% ===============================================

\section{Conclusion}\label{sec:conclusion}

We present a simple method named contrastive-tuning that allows transferring any pre-trained vision model in a zero-shot fashion.
More specifically, the proposed \lit{} setup leads to substantial quality improvements on zero-shot transfer tasks.
It halves the gap between the from-scratch contrastive learning setup, and the per-task supervised fine-tuning setup.
\lit makes it possible to turn publicly available models into zero-shot classifiers using publicly available data, and rival the performance of previous works which rely on more, proprietary data.

We hope that this work motivates future research on how to smartly re-use and adapt already pre-trained models for different research problems.

\paragraph{Acknowledgements} 
We thank Matthias Minderer and Josip Djolonga for help on robustness evaluations; Chao Jia and Zhen Li for discussions on the image-text dataset; Ting Chen for feedback on the initial version of the paper; Jordi Pont-Tuset for help on the image-text retrieval evaluation; Jeremiah Harmsen for inspirations on the title; Jakob Uszkoreit for discussions on data augmentations; Krishna Srinivasan for discussions on the Wikipedia based image text dataset; Beer Changpinyo for discussions on conceptual captions dataset; Maxim Neumann for help on zero-shot eval and T5 text models; the Google Brain team at large for providing a supportive research environment.

%%%%%%%%% REFERENCES
{\small
\bibliographystyle{ieee_fullname}
\bibliography{c}
}

%%%%%%%%% APPENDIX
\clearpage
\appendix

% ===============================================
% ===============================================

\section{Is this specific to ViT image models?}
\label{sec:more_image_architectures}

No.
In the main paper, we only used ViT models for all experiments. Could it be that \lit only works with ViT models, or is in some way specific to the Transformer architecture?

In order to verify that this is not the case, we applied the same recipe to comparably-sized models of different families. Table~\ref{table:more_image_architectures} shows the zero-shot performance with \lit on the CC12M dataset for ViT~\cite{vit}, Mixer~\cite{mixer}, and ResNet~\cite{bit}; all pre-trained on ImageNet21k. Following~\cite{efficiency_misnomer}, we report parameter count, inference speed, and FLOPs to indicate our attempt to match the “model size”. The results show that \lit works for different model families, but also confirm the finding of~\cite{clip} that ViT models do seem more amenable to learning image-text mappings than other architectures of similar size.

\begin{table}[h]
  \newcolumntype{C}{>{\centering\arraybackslash}X}
  \newcolumntype{R}{>{\raggedleft\arraybackslash}X}
  \setlength{\tabcolsep}{0pt}
  \setlength{\extrarowheight}{5pt}
  \renewcommand{\arraystretch}{0.75}
  \centering
  % Format is type{width}, p: paragraph, top-align, C is custom, c is squeezed.
  \begin{tabularx}{\linewidth}{p{1.7cm}p{0.1cm}Cp{0.1cm}Cp{0.1cm}Cp{0.1cm}Cp{0.1cm}Cp{0.2cm}Cp{0.1cm}C}
    \toprule[1pt]
     \bf{Model} &&
    {\centering \rotatebox{90}{\bf{0shot}}} &&
    {\centering \rotatebox{90}{\bf{Adapt}}} &&
    {\centering \rotatebox{90}{\bf{I$\rightarrow$T}}} &&
    {\centering \rotatebox{90}{\bf{T$\rightarrow$I}}} &&
    {\centering \rotatebox{90}{\bf{Param}}} &&
    {\centering \rotatebox{90}{\bf{Speed}}} &&
    {\centering \rotatebox{90}{\bf{FLOPs}}} \\
    \midrule
   ViT-B/32 &&  60.7 &&  79.1 &&  41.3 &&  25.0 && 197\,M &&  2855 && 12\,G \\
 Mixer-B/32 &&  57.1 &&  75.9 &&  37.5 &&  22.9 && 169\,M &&  4208 && \phantom{0}9\,G \\
   BiT-M-R50 &&  55.2 &&  77.6 &&  37.3 &&  23.9 && 134\,M &&  2159 && 11\,G \\
    \bottomrule[1pt]
  \end{tabularx}
  \caption{\protect\lit with different model families. Showing zero-shot top-1 accuracy on ImageNet in comparison to fine-tuning (column “Adapt”). Inference “Speed” is in images per second per core.}\label{table:more_image_architectures}
\end{table}

% ===============================================
% ===============================================

\section{Larger model capacity yields better results}
\label{sec:model_capacity_impact}

\begin{figure}[b]
    \centering
    \includegraphics[width=0.48\textwidth]{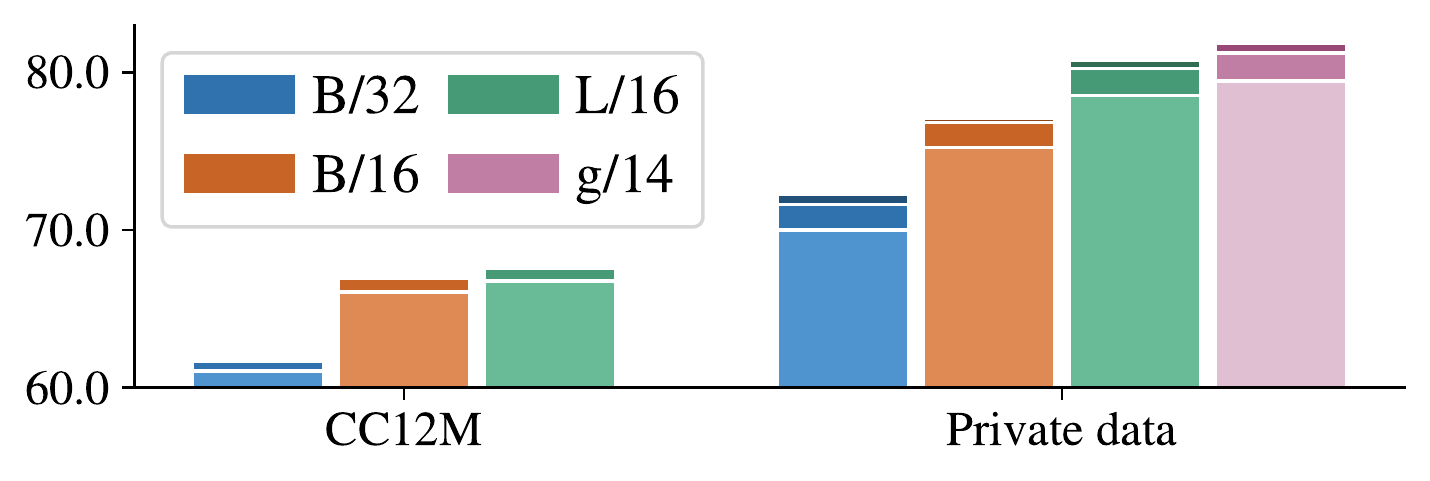}
    \caption{ImageNet zero-shot accuracy [\%] with varying model capacity. Incremental improvemments due to larger \textit{text} towers (base $\rightarrow$ large $\rightarrow$ huge) are shown as stacked bars.}
    \label{fig:model_capacity_impact}
\end{figure}

Increasing the model capacity of the pre-trained image-tower improves zero-shot ImageNet accuracy more than increasing the capacity of the text-tower.
Figure~\ref{fig:model_capacity_impact} shows substantial gains in the private data setup when the image tower capacity is increased from B/32 and base text tower (74.5\%) to g/14 and huge text tower (81.2\%). We take the pre-trained image towers from~\cite{vitg}, and the text towers were trained from scratch.

The improvements in the public CC12M data setup range from 61.1\% with a B/32 image tower and base text tower up to 67.6\% with the L/16 model combined with a large text tower. In this setup, we used pre-trained BERT text towers~\cite{bert} and pre-trained image models from~\cite{augreg} (using the “recommended checkpoints”). Note that in this case the increase from B/16 to L/16 is more modest (from 66.9\% to 67.6\% with the large text tower), and we see a similar improvement in ImageNet zero-shot performance when increasing the text tower size. 

% ===============================================
% ===============================================

\section{Tuning details on our dataset}
\label{appendix:our_details}
We use the pre-trained transformer models from~\cite{vitg}.
ViT-B/32 was used for most of the ablation tests,
and the larger ViT-B/16, ViT-L/16 and ViT-g/14 models are used in Section~\ref{sec:model_capacity_impact} for capacity impact evaluations.
For our best \Lu results, we adopt the ViT-g/14 model pre-trained in~\cite{vitg}.

During contrastive-tuning, we use the AdaFactor optimizer~\cite{adafactor} following~\cite{vitg}.
We use $0.001$ learning rate, and the default $\beta_1=0.9$ and $\beta_2=0.999$ for AdaFactor optimizer.
We use batch size $16384$ by default, unless otherwise noted.
Input image is simply resized to $224\times224$ resolution (apart from $288\times288$ resolution for “g/14*” model).
No weight decay is used during tuning.
We use cosine learning rate schedule with a linear learning rate warmup of 10k steps.
We train our models for 55k steps by default, which equals to about 900 million seen image-text pairs during tuning. 
For our best runs, we scale up the training schedule to 18 billion seen image-text pairs.
We use 128 TPU cores by default for the above experiments, and 256 TPU cores for our best run with 18 billion seen image-text pairs.

In the \Lu setup, we do not attach the optional linear head on the image tower. 
We observe a very small quality improvement without using the image linear head, thus we remove it for simplicity.

\section{Tuning details on CC12m}
\label{appendix:public_details}

We use pre-trained ViT models from~\cite{augreg} (unless otherwise noted, we used the “recommended checkpoints” from that repository). On the text side, we use BERT-base and BERT-large from~\cite{bert} for most experiments. In section~\ref{sec:more_text_models} we use T5-base from~\cite{T5} and mT5-base from~\cite{mt5}.

We use the Adam optimizer ($\beta_1=0.9$, $\beta_2=0.999$) for all models, except for models with Large text tower that were trained with a modified version of AdaFactor from~\cite{vitg} (same settings as described in Section~\ref{appendix:our_details}).
The learning rate is set to $0.001$, and the weight decay to $0.0001$ (using “decoupled” weight decay as described in~\cite{decoupled}). Gradients are clipped at global norm 1.

For training, the images are pre-processed by Inception-style cropping~\cite{inception} to a size of 224 pixels. For evaluation, the images are resized to 224 pixels with bi-linear interpolation without cropping.

When tuning on the CC12M dataset, we train for 20 epochs (200 million seen image-text pairs), which corresponds to 12k steps with a batch size of 16384. The first 50k image-text pairs are used as minival validation set.
The learning rate is ramped up linearly for the first 2k steps and then follows a cosine decay.
Unless otherwise noted, we use the \LU setup with a linear head on the text tower only.

% ===============================================
% ===============================================

\section{How to use YFCC100m?}
\label{appendix:yfcc}

This section is an exploratory analysis of the YFCC100m dataset and provides guidance on what is a good setup for \lit.
For each experiment we run, we try three learning-rates (0.001, 0.0008, 0.0003) and two weight-decays (0.0001 and 0.00001) and report the best result, this allows avoiding biasing conclusions due to sub-optimal hyper-parameters.
We perform the exploration using the small ViT-B/32 AugReg~\cite{augreg} image tower and a BERT base~\cite{bert} text tower and run tuning for 60\,000 steps, although the same conclusions and similar scores are already reachable after 30\,000 steps of tuning.

\begin{figure}[t]
    \centering
    \includegraphics[width=\linewidth]{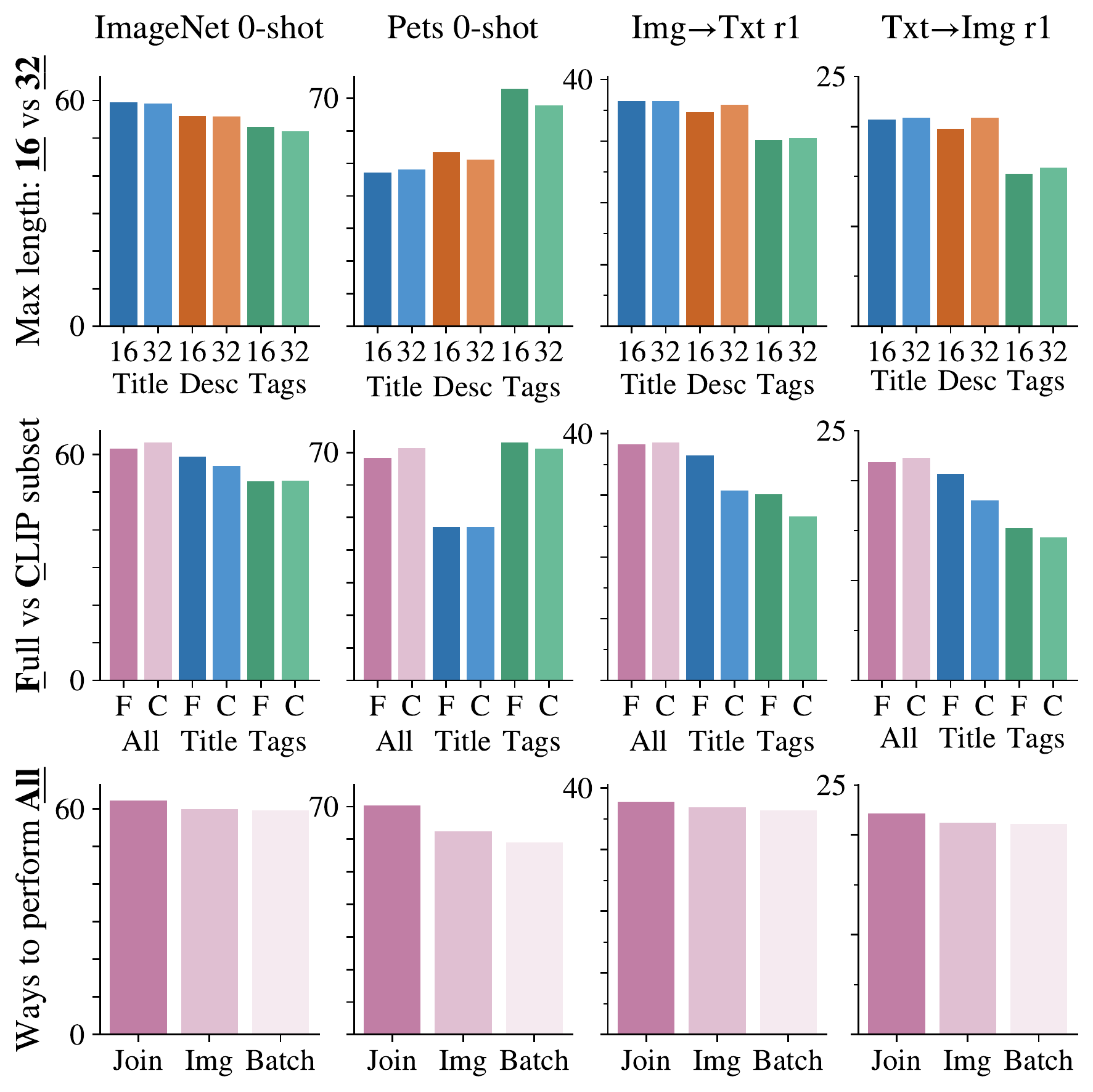}
    \caption{Ablations for YFCC100m.
    \textbf{Top:} even though the description field can be long, the potential benefit of using more than 16 tokens does not outweigh the increased memory and computation cost.
    \textbf{Middle:} When using all text signals, sticking to the CLIP subset is better according to the standard benchmarks, however see also Section~\ref{sec:multilingual}.
    \textbf{Bottom:} Using all three text signals simultaneously for all examples works better than sampling one per image or per batch.}
    \label{fig:yfcc}
\end{figure}

The YFCC100m dataset comes with a rich set of annotations for each image, including camera settings and geolocation.
Out of all the annotations, three of them are potential candidates for learning image-text pairings: the image's title, a description, and a set of free-form tags.
However, only partially overlapping subsets of 60\,M, 30\,M, and 65\,M images come with a title, description, or tags, respectively.
We first explore which supervision signal is most useful.
For the description, we simply tokenize the provided text; for the title, we perform basic filtering and remove titles that start with {\tt DSC}, {\tt IMG}, {\tt Picture}, consist of only the word {\tt image} or consist of more than half digits; for the tags, we randomly shuffle their order, and join them with a random space, newline, or basic punctuation character in order to get a string which we then tokenize.
The texts vary dramatically in length, we thus try maximum sequence lengths of 16 and 32 tokens.
The first row of Figure~\ref{fig:yfcc} shows the result of this experiment.
The difference between a maximum sequence length of 16 and 32 is small, however the memory savings are substantial and we thus restrict the sequence length to 16 tokens in all further experiments.

In terms of supervision signal, there is no single clear winner. We thus explore three ways of learning from all signals and so also make use of the full 100\,M images. We can either \emph{join}tly optimize them by summing up three contrastive losses for each image, or we can randomly sample one of the three sources for each \emph{image} or for a whole mini\emph{batch}. As can be seen in the bottom row of Figure~\ref{fig:yfcc}, jointly using all signals consistently works better, although it requires triple the amount of passes through the text tower.

Finally, the authors of CLIP~\cite{clip} provide a curated subset of roughly 15\,M images, which contain high quality annotations in English. We refer to this subset as YFCC\textsubscript{CLIP}.
In the middle row of Figure~\ref{fig:yfcc}, we compare how using the \emph{F}ull YFCC100m for \lit compares to using the \emph{C}LIP subset of it. Both seem to perform roughly on par for all signals for classification, but when using only titles or tags and performing image-text retrieval, it is better to apply \lit on the full YFCC100m dataset.

Overall, we obtain the best results with \lit using all text signals jointly on the YFCC\textsubscript{CLIP} subset. However, this investigation was performed with the small ViT-B/32 model, it is likely that a larger model may perform better when using the full dataset.

\section{Effective batch size for contrastive loss}
\label{appendix:batch_size}

\begin{figure}[t]
    \centering
    \includegraphics[width=\columnwidth]{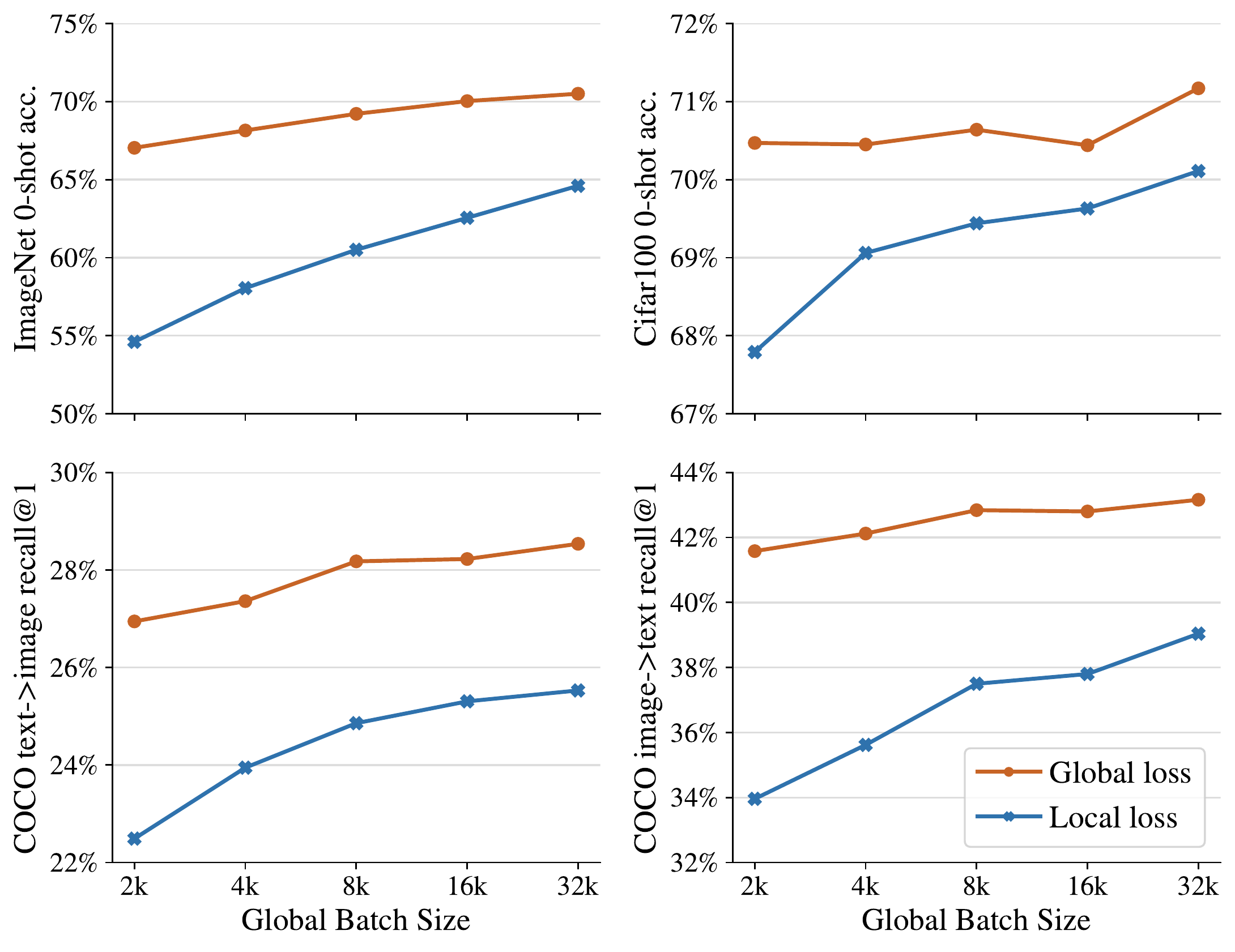}
    \caption{Impact of batch sizes for contrastive loss, including both global contrastive loss and local contrastive loss.}
    \label{fig:batch_size}
\end{figure}

In this section, we study the impact of the effective batch size for contrastive loss.
We use the \Lu setup with a pre-trained B/32 image model, tuned for 900 million seen image-text pairs.
In Figure~\ref{fig:batch_size}, we see a clear improvement when using global contrastive loss.
It has increased the effective batch size for contrastive learning, thus introducing more hard negatives and improving model quality.
Interestingly, we found that larger batch size leads to better performance consistently.
We leave extremely large batch size exploration to future work.

% ===============================================
% ===============================================

\section{Pre-computation for locked image models}
\label{appendix:precompute}

In \lit{} method, the locked image model generates identical embeddings given the same image. 
Based on this characteristic, we use pre-computed image embeddings during tuning.
It allows faster iterations and fitting larger text models in memory, as the image representations are extracted only once and no image models are loaded. 

\begin{figure}[t]
    \centering
    \includegraphics[width=0.236\textwidth]{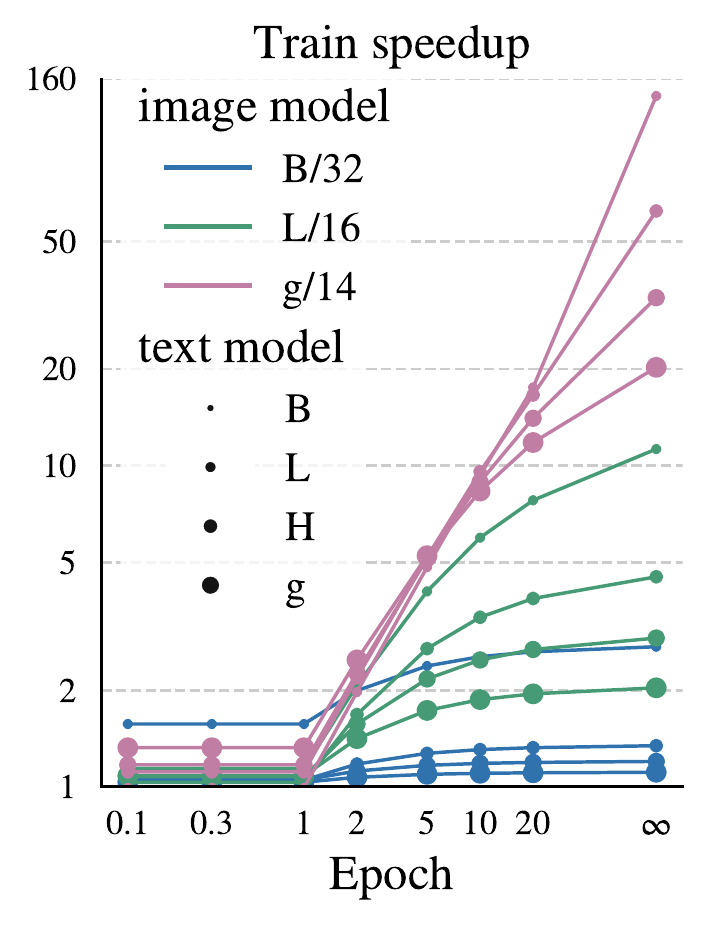}
    \includegraphics[width=0.236\textwidth]{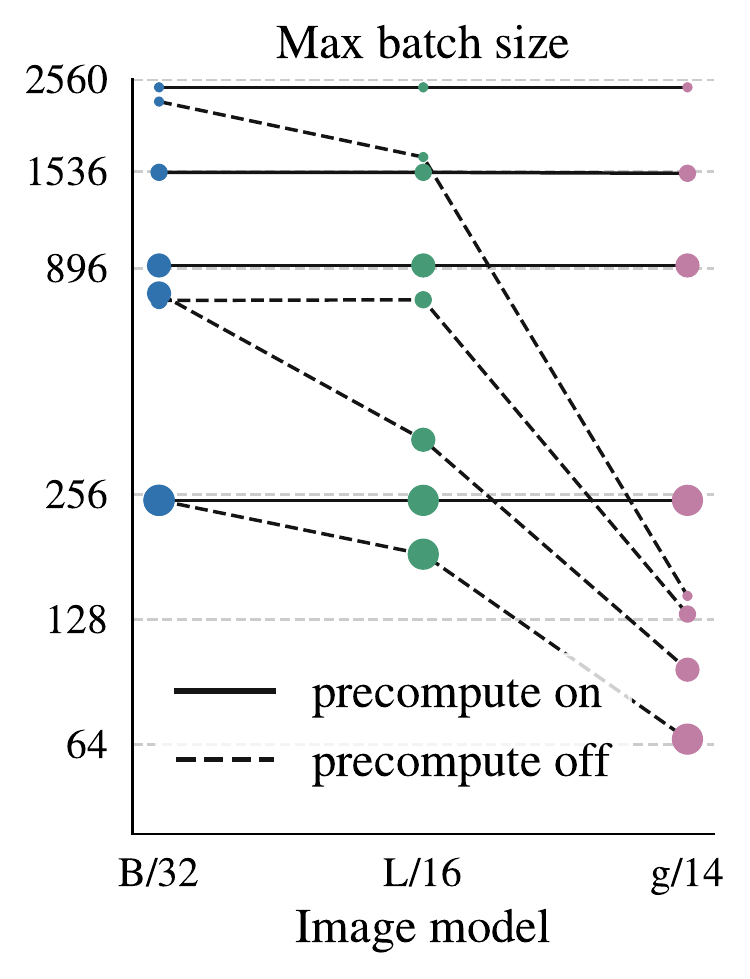}
    \caption{\textbf{Left}: Pre-computing image embeddings accelerates \lit{}, when tuning for more than a single epoch. 
    \textbf{Right}: Pre-computing image embeddings in \lit{} allows larger batch size in memory.}
    \label{fig:precompute}
\end{figure}

Figure~\ref{fig:precompute} left shows how training speeds up as the number of epochs grows.
When training no more than a single epoch, pre-computation keeps a constant speed ratio over re-computation, which increases from one (same speed) to larger than one (speedup) as image model size grows.
After one epoch, pre-computation clearly accelerates training due to reused image representations. 
The speedup ratio becomes more visible as either the number of epochs or the image model size grows.

For experiments with pre-computed image embeddings, we count both pre-computation inference cost and tuning cost. 
Pre-computation will be performed on at most a single epoch on the image-text dataset.
In practice, the pre-computed embeddings can be shared across different experiments, as long as the image tower is identical. 
As a result, the actual cost is even lower than our estimation.
For experiments without pre-computed image embeddings, we count the actual contrastive-tuning cost.

Pre-computation eliminates loading the image model to memory during training, thus allowing larger batch sizes for contrastive loss. 
We search maximum batch sizes on each combination of image and text models with and without pre-computation, and show the results in Figure~\ref{fig:precompute} right. We search for the maximum batch size for each model with a unified setup. We report the maximum batch size that the model can fit on 8 TPU v3 cores.

However, if image augmentations are enabled during training, we may not benefit much from pre-computation.
The model sees different augmented images in multiple epochs.
Nevertheless, the memory benefits still hold.
All metric details are in Table~\ref{tab:precompute_details}.

\begin{table}
\centering
\small
\setlength{\aboverulesep}{0pt}
\setlength{\belowrulesep}{0pt}
\begin{tabular}{l@{\hspace{5pt}}l|r@{\hspace{5pt}}r|r@{\hspace{5pt}}r@{\hspace{5pt}}r|r@{\hspace{5pt}}r}
\toprule
\multicolumn{2}{c|}{}&\multicolumn{2}{c|}{}&\multicolumn{3}{c|}{}&\multicolumn{2}{c}{}\\[-6pt]  % vertical padding
\multicolumn{2}{c|}{\bf{Model}}&\multicolumn{2}{c|}{\bf{Param (M)}}&\multicolumn{3}{c|}{\bf{Max speed}}&\multicolumn{2}{c}{\bf{Max batch}}\\[1.5pt]
\arrayrulecolor{lightgray}\midrule[0.25pt]\arrayrulecolor{black}
Image&Text&Pre&Non&Pre&Non&Inf&Pre&Non\\
\hline
B/32&B&105&195&2439&893&3294&2448&2262\\
B/32&L&320&410&924&688&3294&1528&751\\
B/32&H&640&730&468&390&3294&912&781\\
B/32&g&1007&1097&242&218&3294&248&248\\
L/16&B&105&406&2423&215&273&2448&1663\\
L/16&L&320&621&920&204&273&1528&754\\
L/16&H&640&942&465&160&273&912&347\\
L/16&g&1007&1308&240&118&273&248&184\\
g/14&B&105&1094&2409&17&17&2448&146\\
g/14&L&320&1310&932&15&17&1520&132\\
g/14&H&641&1630&467&14&17&912&97\\
g/14&g&1008&1997&243&12&17&248&66\\
\bottomrule
\end{tabular}
\caption{Pre-computation details. \emph{Max speed} and \emph{Max batch} describe metrics collected by maximum speed (img/sec/core) and batch size, respectively, corresponding to Figure~\ref{fig:precompute}. \emph{Pre} and \emph{Non} are metrics with and without pre-computation respectively; \emph{Inf} describes pre-computation inference speed, which is only affected by image models. All experiments are run on 8 TPU v3 cores.}
\label{tab:precompute_details}
\end{table}

\section{Learning rate schedules}
\label{appendix:lr_schedules}

\begin{figure}[t]
    \centering
    \includegraphics[width=.35\textwidth]{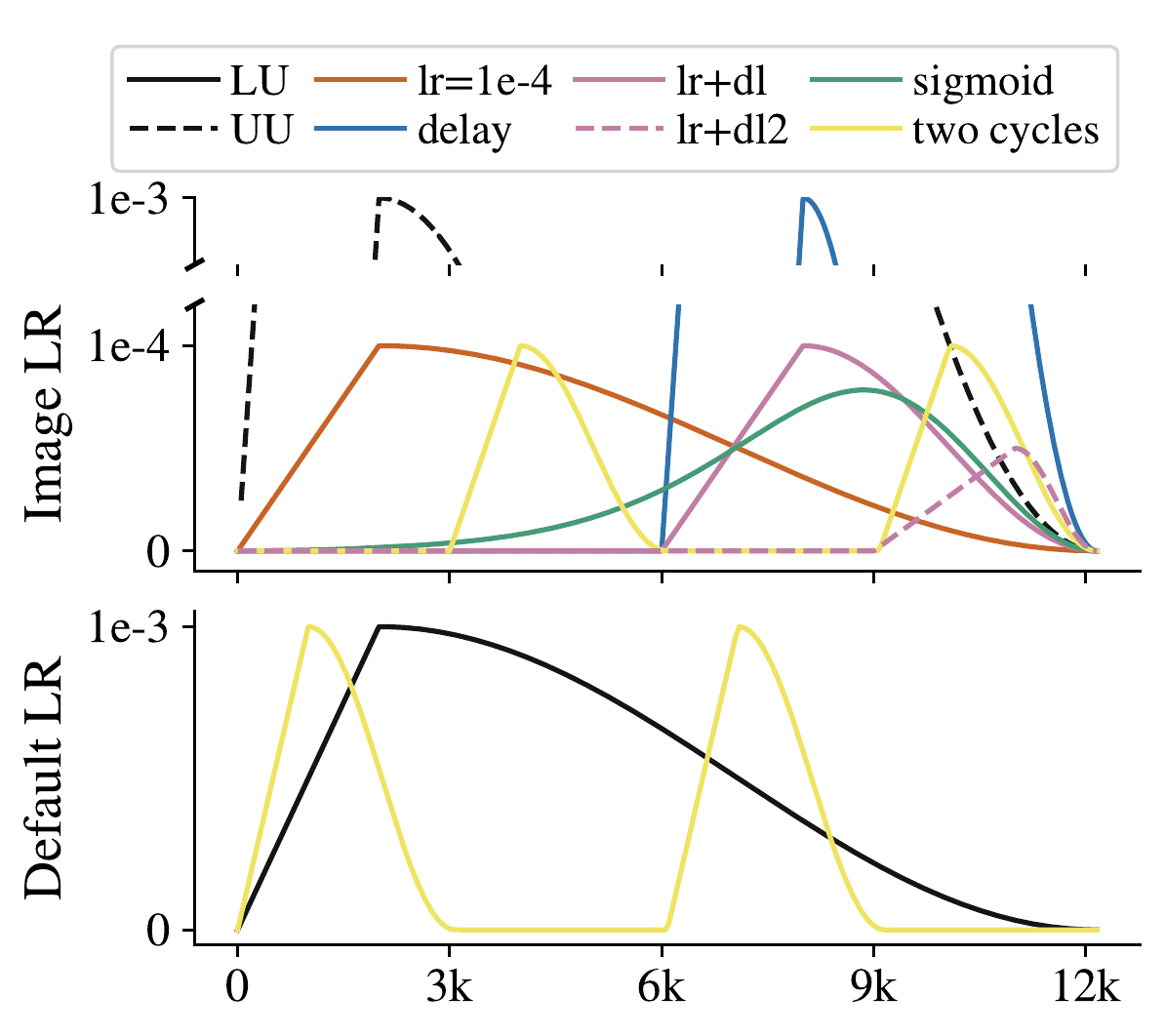}
    \caption{Different learning rate schedules. Note that the default LR schedule is shown in black in the lower part of the figure.}
    \label{fig:lr_schedules}
\end{figure}

\begin{figure}[t]
    \centering
    \includegraphics[width=.35\textwidth]{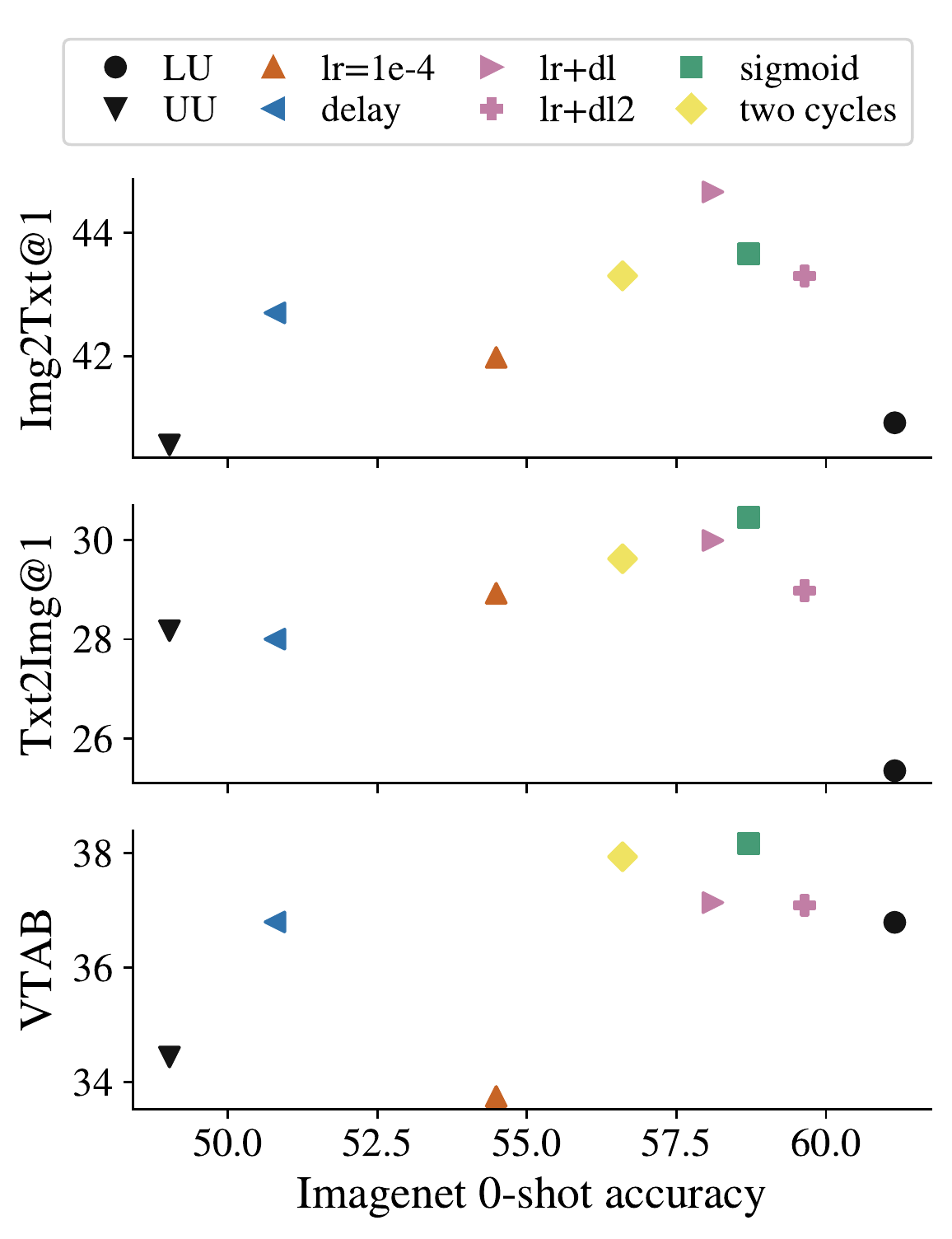}
    \caption{ITR and VTAB metrics as a function of ImageNet 0-shot accuracy for different LR schedules.}
    \label{fig:lr_pointcloud}
\end{figure}

For most of the experiments, weights were either completely locked, or trained with the same learning rate schedule (linear warmup and cosine decay). We experimented with different learning rate schedules (Figure~\ref{fig:lr_schedules}), mainly varying how the image tower was updated. We observed that training the image tower with a smaller learning rate and/or delaying training of the image tower resulted in better retrieval metrics (Figure~\ref{fig:lr_pointcloud}).

The default schedules (LU and UU) have the best and worst ImageNet 0-shot accuracy of all tried learning rate schedules. Compared to UU, both ITR/VTAB metrics and ImageNet 0-shot accuracy improve modestly, when the image learning rate is only scheduled for the second half of the training (“delay"). The ImageNet 0-shot accuracy improves more but the VTAB accuracy drops when the learning rate is set to a smaller value (“lr=1e-4"). Combining the delay with the smaller learning rate (“lr+dl") further improves both ITR/VTAB metrics and ImageNet 0-shot accuracy. A similar result is achieved by multiplying the learning rate in the UU setting with a sigmoid function (“sigmoid"). Alternating between freezing image tower and rext tower (“two cycles") finally performs somewhere between “lr+dl" and “lr=1e-4" schedules.

\section{Zero-shot transfer details}
\label{appendix:zero-shot-details}
\subsection{Classification}

We follow CLIP~\cite{clip} for the zero-shot transfer evaluation. We use the identical ImageNet class label names and the same 80 prompt templates as in CLIP.
During evaluation of private LiT models, we first resize the test image and then central crop with 0.875 aspect ratio to the target resolution. More specifically, we use $224\times224$ target resolution for CIFAR dataset and $288\times288$ target resolution for the remaining datasets.
For all the public LiT models, we resize all test images to $224\times224$ for simplicity.

\subsection{VTAB Evaluation}
\label{appendix:vtab}
The Visual Task Adaptation benchmark\cite{vtab} consists of 19 diverse visual tasks. We refer readers to the original publication for details about each dataset; here we just mention that they are split into three categories:
\begin{itemize}
    \item \textbf{Natural}: These tasks contain classical “natural” real-world images obtained with a camera, such as vehicles, pets, scenery and household objects.
    \item \textbf{Specialized}: These are datasets of arguably “natural” images which were captured with specialised photographic equipment, such as satellite photographs and medical images.
    \item \textbf{Structured}: These assess understanding of scenes structure in some way, predominately from synthetic environments. Example tasks include 3D depth estimation and counting.
\end{itemize}
Note that there is significant overlap with the datasets assessed in~\cite{clip}, but it is not guaranteed that the same data splits were used.

\textbf{Evaluation protocol.} 
Previous works\cite{clip} define task-specific prompts and class names, but it is not clear exactly how an optimal set of prompts for a given task was chosen.

For VTAB, we define a search space of image preprocessing, prompt templates and classes, where the latter two are often per-task (e.g. using `a` `satellite` `photo` `of` `...` or `an` `overhead` `photo` `of` `...` for tasks involving satellite imagery). All such settings are tried on a small validation set of 800 images, and the optimal setting is then run on the official VTAB test set.

We note this is arguably not \textit{zero-shot} transfer, but believe it is a principled and reproducible approach.

\textbf{Prompts used.}
For all tasks, we considered 3 default sets of prompts
\begin{enumerate}
\item \texttt{A photo of a \cls}
\item \texttt{\cls}
\item The 6 CLIP prompts used for ImageNet\footnote{https://github.com/openai/CLIP/blob/main/data/prompts.md}
\end{enumerate}

We also consider some task specific prompts/class name settings. Note that these two degrees of freedom are orthogonal, and a text setting is defined by both. They are shown in  Table~\ref{tab:vtab_prompts_default} and Table~\ref{tab:vtab_prompts_custom}. Not all of these prompts were equally useful; some are redundant, providing equal performance gain as other settings, and some do not provide performance gains at all. We show the performance delta comparing only the default prompts versus including a given text variant as well, to give a rough idea of how beneficial it was.

\begin{figure}[t]
    \centering
    \includegraphics[width=.7\linewidth]{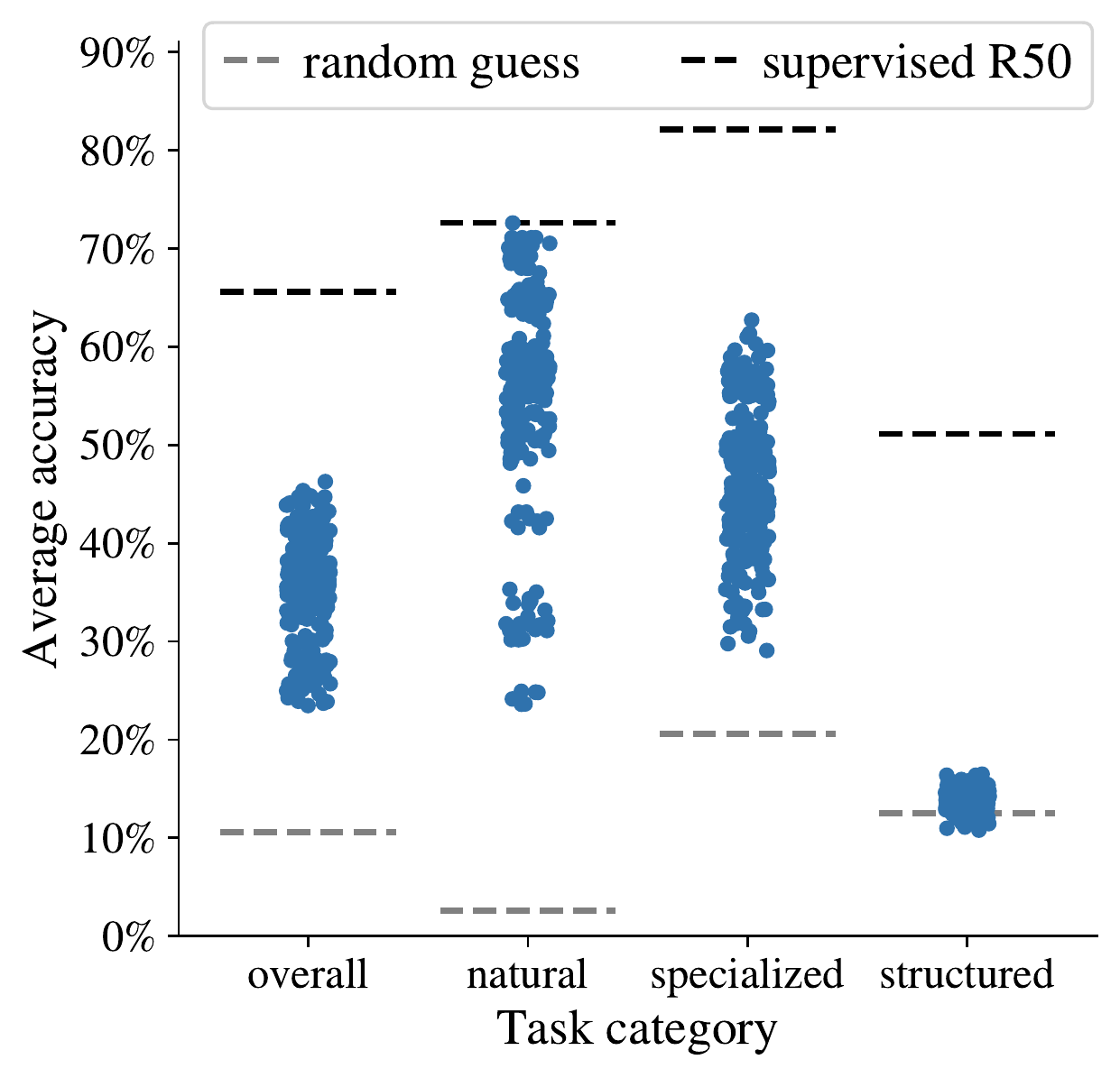}
    \caption{Performance of zero-shot classification models across different VTAB categories. Each dot is a zero-shot model evaluation.}
    \label{fig:vtab_cat}
\end{figure}

\textbf{Can we assess zero-shot performance using VTAB?}
The strength of such a diverse benchmark is in the variety of its label spaces. ImageNet classes, though very fine-grained, are fairly generic. However, VTAB also includes \textit{structured tasks} which are designed to assess the model's competence at tasks which aren't object recognition, such as counting and assessing distances and angles.
This presents interesting difficulties for solving in a zero-shot natural language grounded manner. Figure \ref{fig:vtab_cat} shows the zero-shot performance of many models developed for this paper. Their detailed performance is not important here - the gray lines show what a “random guesser” would achieve on each VTAB category. It is not an obvious number, as performance across categories is an average of all the constituent datasets, which have varying numbers of classes.
It is clear from this figure that the structured performance does not significantly deviate from random guessing, despite extensive efforts in prompt engineering. We leave it as an open - and very interesting - research direction to figure how to make such models count and assess distances. Furthermore, though contrastive image-text training on the web can largely match supervised models on natural tasks, further improvements are needed on more specialist tasks.

\subsection{Cross-modal retrieval}

We compute retrieval metrics on MSCOCO captions~\cite{mscoco}, reporting the numbers on the test set (5\,000 images, 25\,010 captions).
For the image to text retrieval, we rank all texts by decreasing cosine similarity of their embedding with the image embedding, and then report the fraction of images that ranks the correct text within the first (1, 5, 10) positions as the Recall@1, Recall@5, Recall@10 metrics.
For the text to image retrieval, we compute the same metric, but ranking images and averaging over all texts.
When showing a single number, we always refer to the Recall@1 metric.

\section{Multilingual details and limitations}
\label{appendix:multilingual_details}

\begin{figure*}[t]
    \centering
    \includegraphics[width=\textwidth]{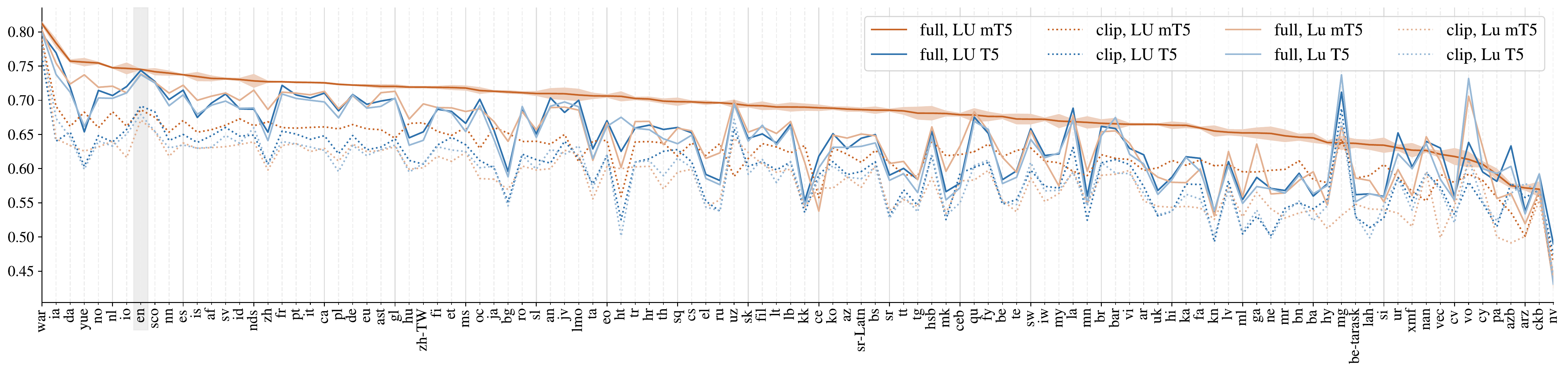}
    \caption{Fully detailed evaluation of the multilingual models on WIT.}
    \label{fig:xlang_wit_wide}
\end{figure*}

\textbf{Extra results.}
Table \ref{tab:yfcc_xlang} shows the English zero-shot ImageNet classification performance of different English and multilingual T5 models, with \lit{} on YFCC\textsubscript{CLIP} vs. YFCC100m. We note that training on the larger, more diverse, multilingual set does not come at the expense of English performance.
\begin{table}
\centering
\begin{tabular}{>{\kern-\tabcolsep}lcccccc<{\kern-\tabcolsep}}
\toprule
 & \multicolumn{3}{c}{CLIP subset} & \multicolumn{3}{c}{Full} \\
{} &    \textbf{ImgNet} &   \textbf{T$\rightarrow$I} &   \textbf{I$\rightarrow$T} & \textbf{ImgNet} &   \textbf{T$\rightarrow$I} &   \textbf{I$\rightarrow$T} \\
\midrule
T5 &        58.9 &  14.5 &  22.6 &     62.4 &  19.6 &  34.3 \\
\rowcolor{gray!20} 
\textit{ + pt} &        58.5 &  17.2 &  29.1 &     62.3 &  \textbf{20.1} &  \textbf{34.5} \\
mT5       &        58.7 &  14.4 &  23.1 &     62.1 &  18.5 &  32.6 \\
\rowcolor{gray!20}
\textit{ + pt} &        58.4 &  15.6 &  25.1 &     \textbf{62.6} &  18.9 &  33.6 \\
\bottomrule\end{tabular}
\caption{Training on the full YFCC100m data significantly improves all metrics compared to the CLIP subset. Gray rows are with text pre-training.}
\label{tab:yfcc_xlang}
\end{table}

\textbf{Wiki-Image Text as an evaluation benchmark.} We noted qualitatively that, as one may expect from Wikipedia, a large proportion of examples are about entities such as people, places, or art. When translated to other languages, proper nouns are usually kept as is - especially if the two languages share an alphabet. This makes it an imperfect dataset to benchmark multilinguism as monolingual models will score higher than they should.

\textbf{Tokenization subtleties.} The sentencepiece tokenizers, when faced with unknown vocabulary, will default to byte encoding. This is not a perfect catch-all; in such circumstances models cannot take advantage of pre-training, and the resultantly very long sequences will not fit in the 16-token maximum length used in this paper. It is nevertheless better than the `[UNK]` tokens produced by BERT's WordPiece tokenizer; with SentencePiece, even with an imperfect vocabulary, the model has a chance to adapt. This explains why even with an ill-suited English-only vocabulary, the T5 models can still learn decent representations of non-English languages.

\textbf{Translation of prompts.} One obvious factor worth noting is that, in our setup, non-English languages may be impacted by imperfect translations. This likely means non-English performance is underestimated.

More subtly, we note that many languages - especially those with Latin alphabets - often use the English word for very niche or specific items. For example, at the time of writing, the Vietnamese translation of `I` `took` `a` `photo` `of` `an` `airship` contains the word `airship` verbatim. The contrastive model can in principle pick out the word `airship`, ignore all the Vietnamese, and retain decent performance despite not understanding Vietnamese at all.

\begin{figure}[t]
    \centering
    \includegraphics[width=.8\linewidth]{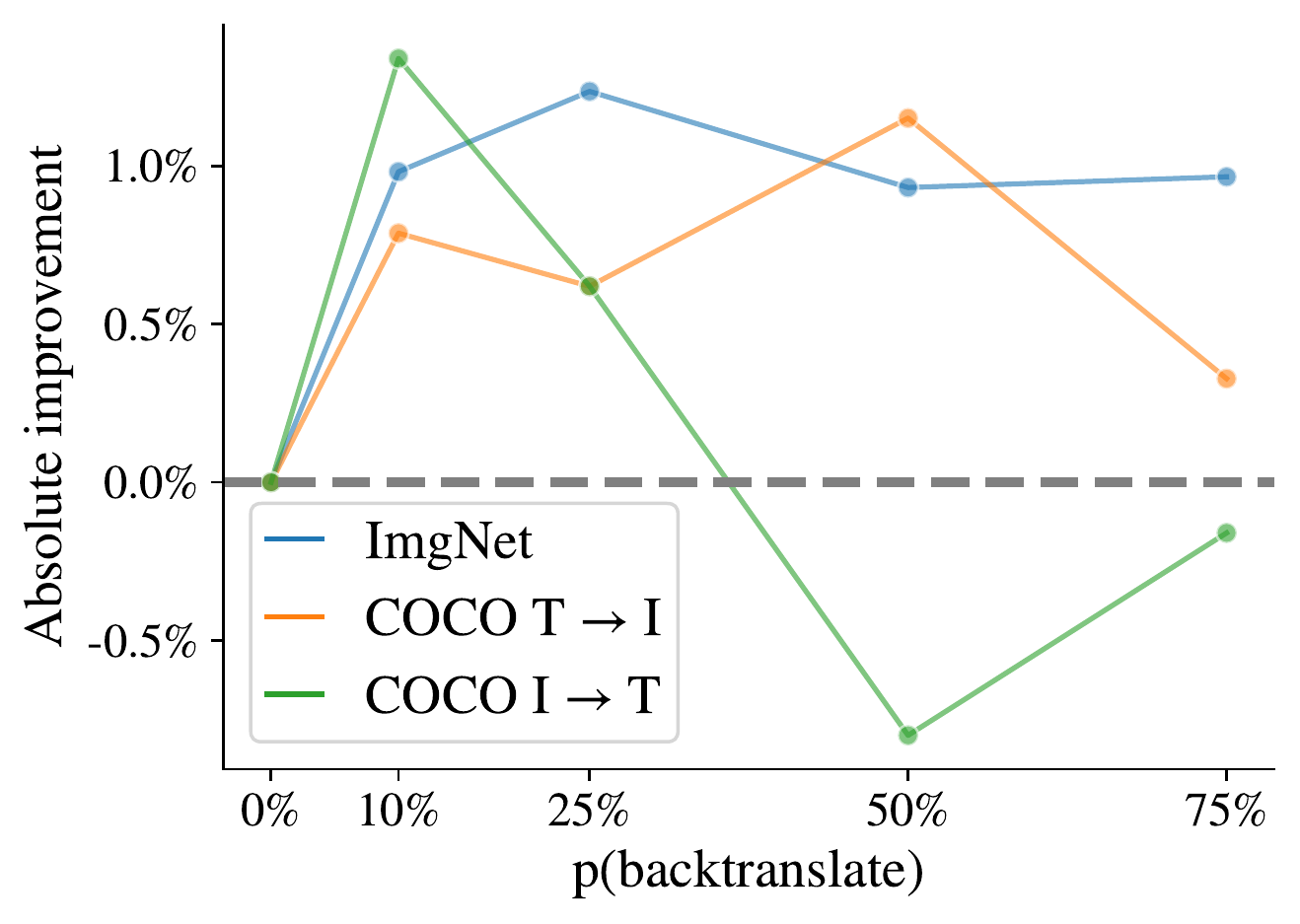}
    \caption{Backtranslating data as a form of data augmentation improves performance across most metrics.}
    \label{fig:backtranslate}
\end{figure}

\textbf{Backtranslation as data augmentation.}
Backtranslation\cite{backtranslate} - translating to a language and back again, in order to generate slightly different versions of a given text - is a common augmentation in NLP. We run some experiments to see whether it works for contrastive image-text training.
We again use an online translation service to translate the texts in CC12M to and from 9 different languages. This probability is shared across the languages i.e. a backtranslation probability of 0.5 with 5 different backtranslate candidates means there is a 50\% chance of picking the original ground truth and a 10\% chance each of picking one of the backtranslated candidates.
Figure \ref{fig:backtranslate} shows the effect of this augmentation on \lit{} using an AugReg ImageNet21k pre-trained ViT-B/16 model. Backtranslation is fairly useful up to certain point, with 10\% giving a good trade-off which improves all metrics.

\section{More de-duplication results}
\label{appendix:more_dedup}

We present more ablation test results using larger architectures. 
We aim to check whether larger architectures be\-ne\-fit more from duplicates, while small architectures do not have enough capacity to overfit to the duplications.
More specifically, we adopt the \Lu setup with a pre-trained ViT-L/16 image model~\cite{vitg}, and from-scratch L size text model.
Table~\ref{table:duplicates_l16} shows the experimental results. We find that the conclusions are consistent with the runs using the ViT-B/32 image model discussed in Section~\ref{sec:deduplication}. 
This is further evidence suggesting that duplications are not the root cause for good zero-shot transfer results.

\begin{table}[t]
  % \small
  \newcolumntype{C}{>{\centering\arraybackslash}X}
  \newcolumntype{R}{>{\raggedleft\arraybackslash}X}
  \setlength{\tabcolsep}{0pt}
  \setlength{\extrarowheight}{5pt}
  \renewcommand{\arraystretch}{0.75}
  \centering
  % Format is type{width}, p: paragraph, top-align, C is custom, c is squeezed.
  \begin{tabularx}{\linewidth}{p{1.5cm}p{0.1cm}Cp{0.1cm}Cp{0.1cm}Cp{0.1cm}Cp{0.1cm}C}
    \toprule[1pt]
     \bf{Dedup} && \bf{\# up.} && \bf{\# down.} && \bf{ImgNet} && \bf{I$\rightarrow$T} && \bf{T$\rightarrow$I}\\
    \midrule
     - && 0 && 0 && 80.2 && 50.4 && 34.6 \\
     test && 2.6M && 76K && 80.2 && 49.0 && 34.3 \\  % 2_575_200
     train+test && 3.6M && 220K && 80.0 && 49.6 && 34.6 \\  % 3_637_630
    \bottomrule[1pt]
  \end{tabularx}
  \caption{Results on three different de-duplication setups, Lu setup with pre-trained ViT-L/16 image model.}\label{table:duplicates_l16}
\end{table}

\section{Image-text dataset comparison}
\label{appendix:align_vs_our}

\begin{table}[t]
  % \small
  \newcolumntype{C}{>{\centering\arraybackslash}X}
  \newcolumntype{R}{>{\raggedleft\arraybackslash}X}
  \setlength{\tabcolsep}{0pt}
  \setlength{\extrarowheight}{5pt}
  \renewcommand{\arraystretch}{0.75}
  \centering
  % Format is type{width}, p: paragraph, top-align, C is custom, c is squeezed.
  \begin{tabularx}{\linewidth}{p{1.5cm}p{0.1cm}Cp{0.1cm}Cp{0.1cm}Cp{0.1cm}C@{\hspace{-1ex}}}
    \toprule[1pt]
     \bf{Task} && \bf{Pairs Seen} && \bf{our} && \bf{ALIGN} && \bf{Diff.}\\
    \midrule
     ImageNet && 900M && 70.1 && 69.8 && 0.3\\
     ImageNet && 3.6B && 72.0 && 71.5 && 0.5\\
     ImageNet && 7.2B && 72.4 && 71.8 && 0.6\\
     ImageNet && 18B && 72.9 && 72.2 && 0.7\\
    \bottomrule[1pt]
  \end{tabularx}
  \caption{Comparing the ALIGN data with our data, which uses simpler text filters. }\label{table:align_vs_our}
\end{table}

Using simpler text filters for our dataset leads to a larger dataset size compared to the ALIGN dataset:
The ALIGN dataset contains 1.8B image-text pairs, while our data set contains 3.6B image-text pairs. 

In table~\ref{table:align_vs_our}, we show the results from training a baseline ViT-B/32 model on both datasets, with the same schedules. 
We vary the training schedule from 900M seen images, to 18B seen images.
We use 18B images to make sure that the training process is long enough to benefit from a larger dataset.
We find that the difference between the two datasets are small when the model is trained for a short period, \ie less than a single epoch.
As the training becomes longer, the impact of the dataset size becomes more visible.

Overall, the above results indicate that larger dataset with simpler filters slightly outperforms a smaller dataset with more filters.
We leave the thorough exploration of this topic to future work.

\begin{figure*}[t]
    \small
    \centering
    \includegraphics[width=0.495\textwidth]{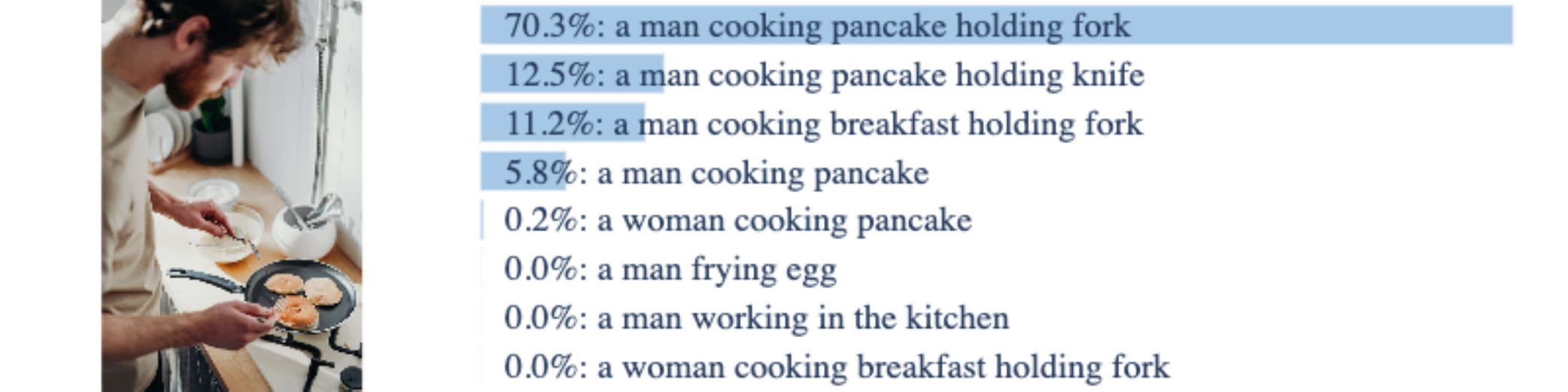}
    \includegraphics[width=0.495\textwidth]{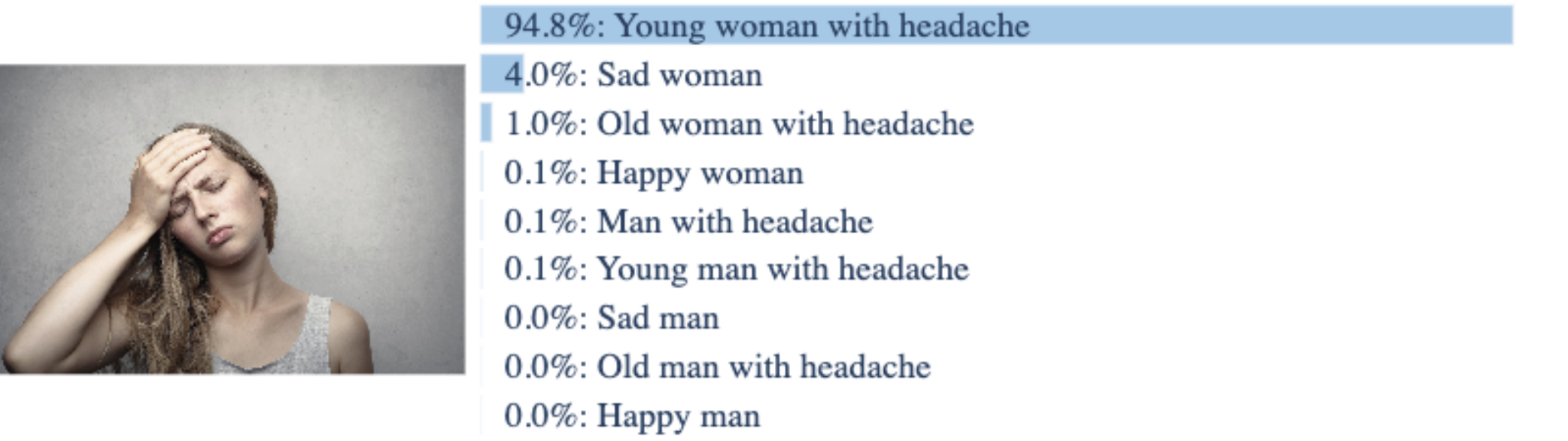}
    \textbf{(a) Nuanced context}: The model can understand information such as actions or implied symptoms.
    \includegraphics[width=0.46\textwidth]{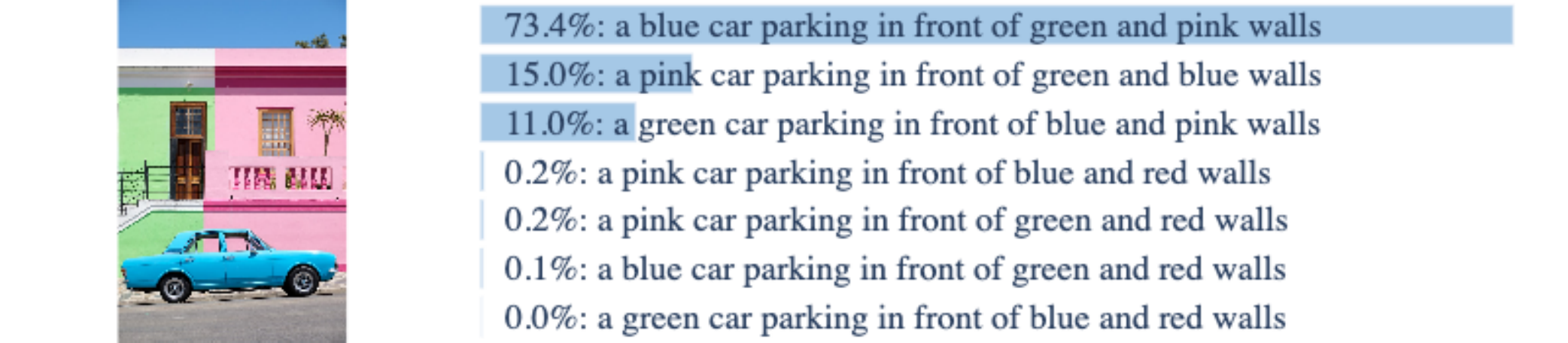}
    \includegraphics[width=0.53\textwidth]{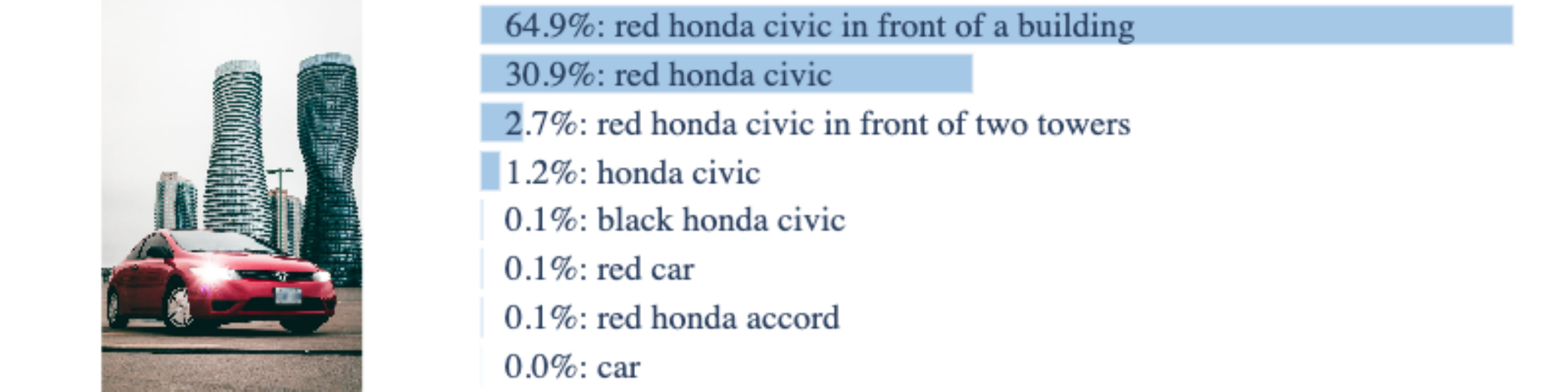}
    \textbf{(b) Richer information}: The model correctly handles colours, background buildings and even car brands.
    \includegraphics[width=0.495\textwidth]{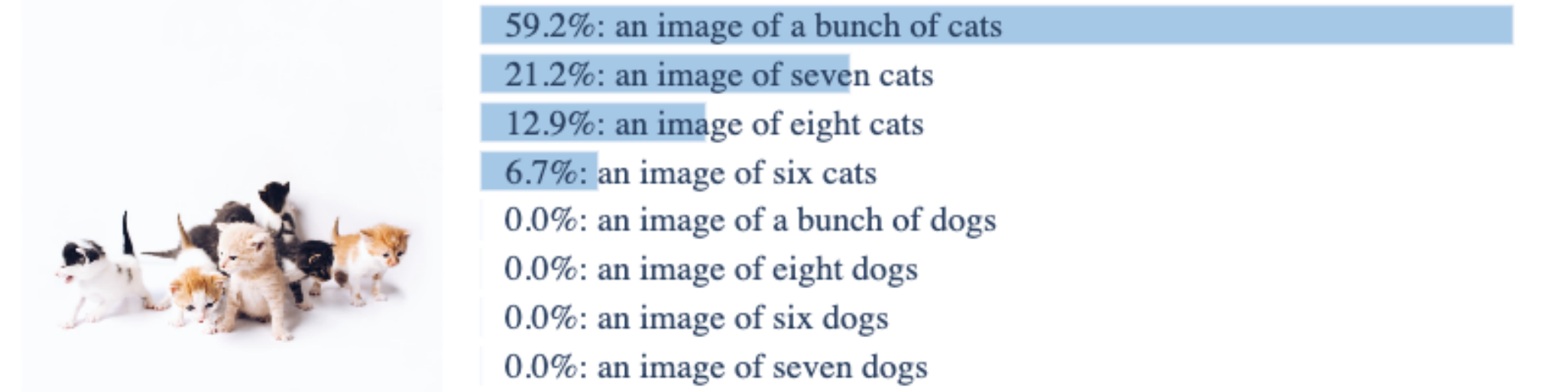}
    \includegraphics[width=0.495\textwidth]{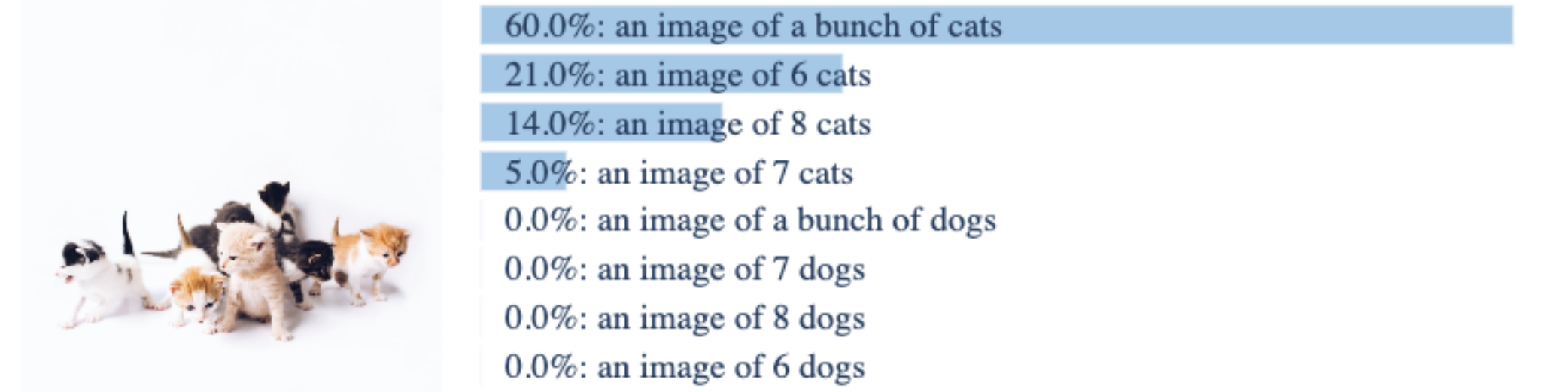}
    \textbf{(c) Counting}: The model does a reasonable job at counting, though prompts like “bunch of cats" are preferred.
    \includegraphics[width=0.495\textwidth]{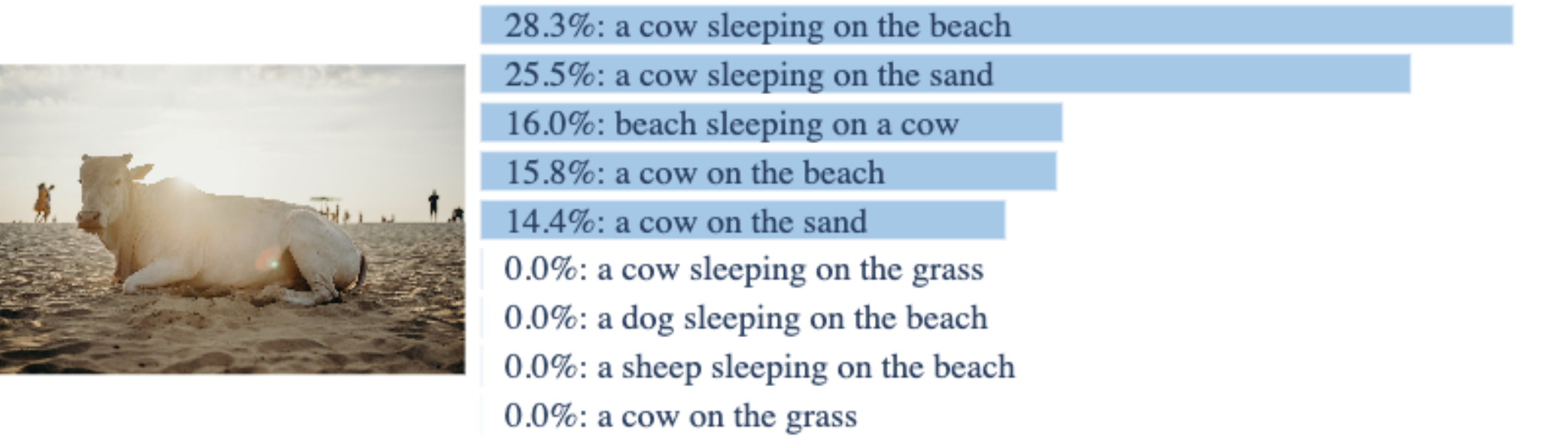}
    \includegraphics[width=0.495\textwidth]{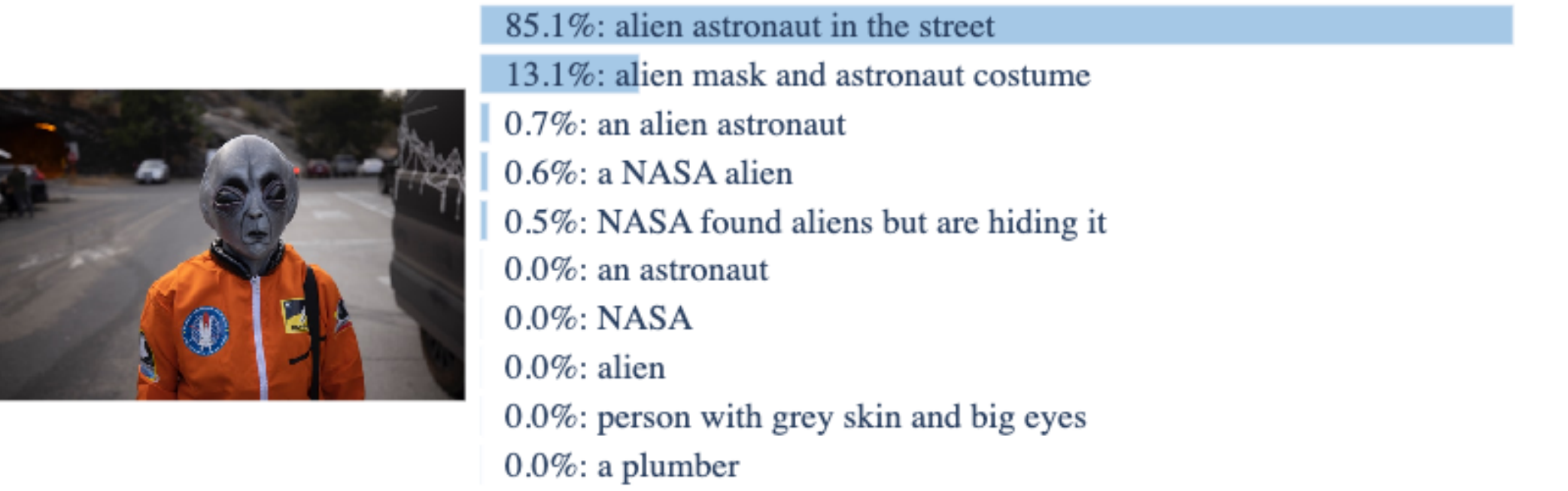}
    \textbf{(d) Esoteric examples}: The model has no problems at identifying rare concepts, like a cow on a beach, or an astronaut alien.
    \caption{Various model predictions.}
    \label{fig:success_examples}
\end{figure*}

\begin{figure*}[h]
    \small
    \centering
    \includegraphics[width=0.495\textwidth]{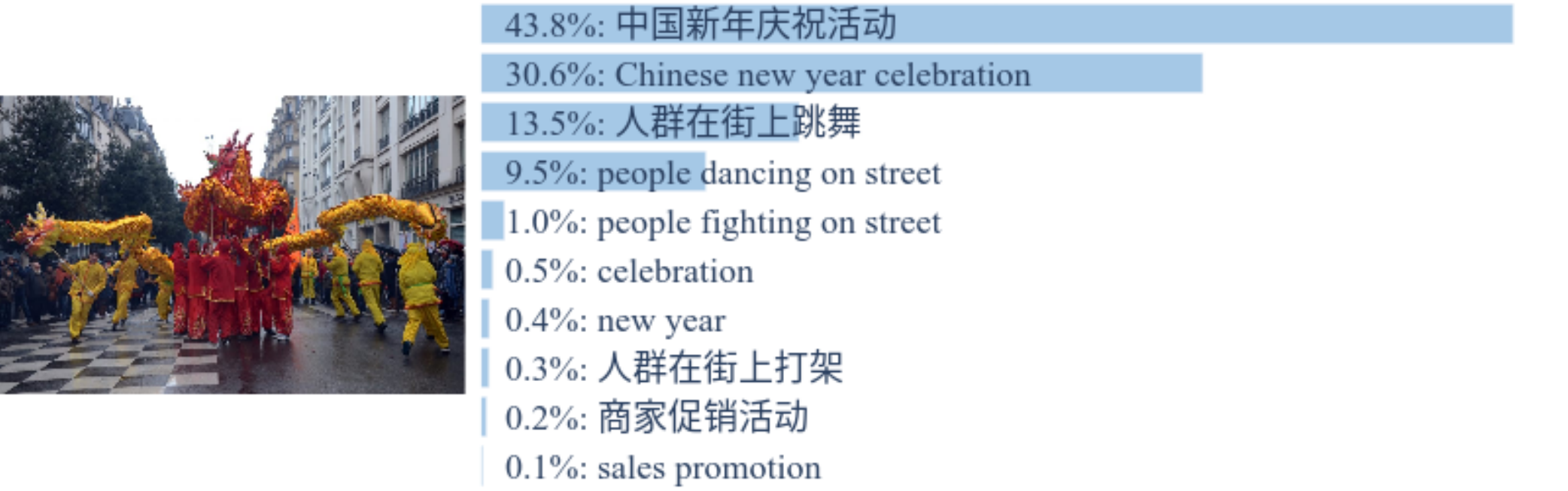}
    \includegraphics[width=0.495\textwidth]{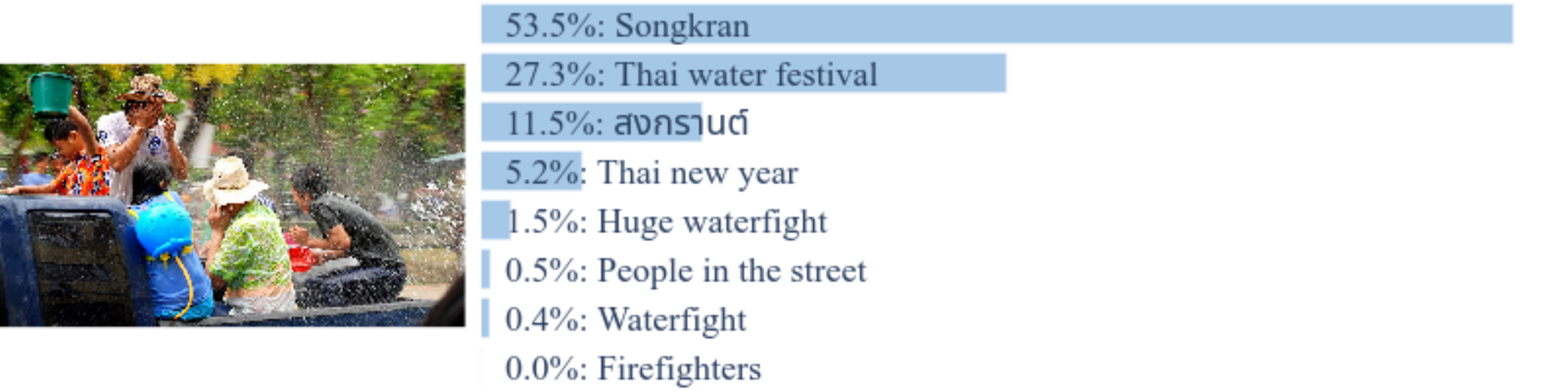}
    \caption{Training on multilingual data allows the model to recognise concepts in multiple languages, including visual concepts which do not directly exist in English.}
    \label{fig:xlang_examples}
\end{figure*}

\begin{figure*}[h]
    \small
    \centering
    \includegraphics[width=0.45\textwidth]{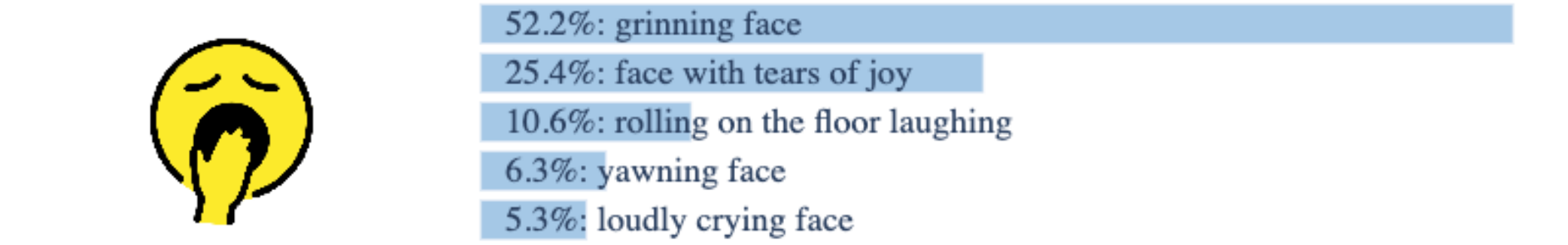}
    \includegraphics[width=0.54\textwidth]{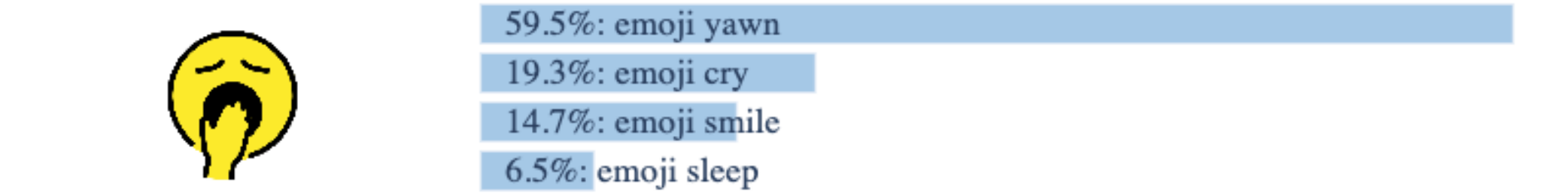}
    \textbf{}
    \caption{Qualitative failures. In the left example, the model ranks the wrong grinning face before the ground truth yawning face. However, by removing the grinning face and adding emoji prompt, the model prefers emoji yawn.}
    \label{fig:failure_examples}
\end{figure*}

\section{Qualitative examples}
\label{appendix:demos}
Though strong classification \& retrieval performance is promising, it arguably probes understanding of very simple concepts. Are \lit{} models really zero-shot learners capable of understanding open vocabularies?

We touch here on a few qualities these models should ideally have, but note that these are not to be considered representative; benchmarks that investigate more than simply fine grained visual classification should be used to more thoroughly understand these phenomena.

\subsection{Private LiT model}
In this section, we present model predictions with manually constructed image-text pairs input. Results from private LiT model are shown in Figure~\ref{fig:success_examples}. We believe that with LiT, we successfully made a pre-trained image model to a zero-shot learner, that supports classification and retrieval with open vocabularies instead of a fixed label set.

\subsection{Multi-lingual model}
Thanks to LiT on the multilingual dataset, the model also supports inputs using different languages. In Figure~\ref{fig:xlang_examples}, we show results both in Thai and Chinese. The model recognized the “Songkran" event in Thai, and the “Chinese Spring Festival" event in Chinese; it nonetheless also ranks English translations or transliterations quite highly, which is likely reflective of the data distribution. 
Multilingual capability makes our models more inclusive and accessible to non-English speakers.

\subsection{Model failures}
We present model failures in Figure~\ref{fig:failure_examples}. We show examples of how one can slightly change the text candidastes to manipulate the model output; one can easily force a desired answer by tuning other text candidates to rank lower.

\begin{table*}[h]
  % \small
  \newcolumntype{C}{>{\centering\arraybackslash}X}
  \newcolumntype{R}{>{\raggedleft\arraybackslash}X}
  \setlength{\tabcolsep}{0pt}
  \setlength{\extrarowheight}{5pt}
  \renewcommand{\arraystretch}{0.75}
  \centering
  % Format is type{width}, p: paragraph, top-align, C is custom, c is squeezed.
  \begin{tabularx}{\linewidth} {p{4cm}p{0cm}Cp{1cm}c}
    \toprule[1pt]
     \bf{Dataset} && \bf{Prompts} && \bf{Delta} \\
    \midrule
     dtd \emph{v3.0.1} && \texttt{a \cls~ texture} && +0.6\% \\
     \arrayrulecolor{black!30}\midrule
     flowers \emph{v2.1.1} && \texttt{a \cls~ flower} && +1.1\% \\
     flowers \emph{v2.1.1} && \texttt{a \cls~ plant} && +0.4\% \\
     \midrule
     pets \emph{v3.2.0} && \texttt{a type of pet \cls~} && +1.0\% \\
     pets \emph{v3.2.0} && \texttt{a \cls~ texture} && +0.4\% \\
     pets \emph{v3.2.0} && \texttt{\cls~, an animal} && +0.7\% \\
     \midrule
     svhn \emph{v3.0.0} && \texttt{the number \cls~} && +3.0\% \\
     svhn \emph{v3.0.0} && \texttt{a street sign with the number \cls~} && +2.8\% \\
     \midrule
     camelyon \emph{v2.0.0} && \texttt{a histopathology slide showing \cls~} && +1.5\% \\
     camelyon \emph{v2.0.0} && \texttt{histopathology image of \cls~} && +0.9\% \\
     \midrule
     eurosat \emph{v2.0.0} && \texttt{a satellite photo of \cls~} && +3.2\% \\
     eurosat \emph{v2.0.0} && \texttt{\cls~ from above} && +2.4\% \\
     eurosat \emph{v2.0.0} && \texttt{an aerial view of \cls~} && +3.3\% \\
     \midrule
     resisc \emph{v3.0.0} && \texttt{a satellite photo of \cls~} && +3.4\% \\
     resisc \emph{v3.0.0} && \texttt{\cls~ from above} && +2.1\% \\
     resisc \emph{v3.0.0} && \texttt{an aerial view of \cls~} && +4.7\% \\
     \midrule
     retino \emph{v3.0.0} && \texttt{a retinal image with \cls~} && +9.7\% \\
     retino \emph{v3.0.0} && \texttt{a retina with \cls~} && +6.3\% \\
     retino \emph{v3.0.0} && \texttt{a fundus image with signs of \cls~} && +6.3\% \\
     \midrule
     clevr-count \emph{v3.1.0} && \texttt{\cls~ objects} && +0.1\% \\
     clevr-count \emph{v3.1.0} && \texttt{\cls~ things} && +0.2\% \\
     clevr-count \emph{v3.1.0} && \texttt{a photo of \cls~ objects} && +0.1\% \\
     \midrule
     dsprites-pos \emph{v2.0.0} && \texttt{an object located \cls~} && +0.0\% \\
     \midrule
     dsprites-orient \emph{v2.0.0} && \texttt{an object rotated at \cls~} && +0.1\% \\
     dsprites-orient \emph{v2.0.0} && \texttt{something rotated at \cls~} && +0.0\% \\
     dsprites-orient \emph{v2.0.0} && \texttt{\cls~ rotation} && +0.0\% \\
     dsprites-orient \emph{v2.0.0} && \texttt{something at a \cls~ angle} && +0.0\% \\
     \midrule
     smallnorb-azmth \emph{v2.0.0} && \texttt{an object rotated at \cls~} && +0.0\% \\
     smallnorb-azmth \emph{v2.0.0} && \texttt{something rotated at \cls~} && +0.0\% \\
     smallnorb-azmth \emph{v2.0.0} && \texttt{\cls~ rotation} && +0.0\% \\
     smallnorb-azmth \emph{v2.0.0} && \texttt{something at a \cls~ angle} && +0.0\% \\
     \midrule
     smallnorb-elev \emph{v2.0.0} && \texttt{an object rotated at \cls~} && +0.0\% \\
     smallnorb-elev \emph{v2.0.0} && \texttt{something rotated at \cls~} && +0.0\% \\
     smallnorb-elev \emph{v2.0.0} && \texttt{\cls~ rotation} && +0.0\% \\
     smallnorb-elev \emph{v2.0.0} && \texttt{something at a \cls~ angle} && +0.0\% \\
    \arrayrulecolor{black!100}\bottomrule[1pt]
  \end{tabularx}
  \caption{\normalsize Prompts swept over for VTAB tasks. Performance deltas are shown as mean test accuracy improvement per-task compared to just using the default three prompts. The default class names from TensorFlow Dataset (TFDS) are used in this table. TFDS versions are given alongside task names.}
  \label{tab:vtab_prompts_default}
\end{table*}

\newpage
\onecolumn
\clearpage
\setlist{nosep,after=\vspace{-\baselineskip}}
\begin{longtable}[h]{p{0.2\textwidth}p{0.58\textwidth}p{0.22\textwidth}}
\toprule
\multicolumn{3}{c}{\textbf{svhn} \emph{v3.0.0}} \\
% svhn variant=number

\midrule
% svhn variant=number_letters

Prompts:
\small
\begin{itemize}[itemsep=1pt,topsep=1pt,leftmargin=12pt]
	\item \texttt{the number \cls~}
\end{itemize}
\normalsize &
Class names:

\small
\begin{enumerate}[itemsep=1pt,topsep=1pt,leftmargin=12pt]
	\item \clsfmt{zero}
	\item \clsfmt{one}
	\item \clsfmt{two}
	\item \clsfmt{three}
	\item \clsfmt{four}
	\item \clsfmt{five}
	\item \clsfmt{six}
	\item \clsfmt{seven}
	\item \clsfmt{eight}
	\item \clsfmt{nine}
\end{enumerate}
\normalsize
&
Delta: \small 
+2.3\%
\normalsize \\

\arrayrulecolor{black!30}\midrule
% svhn variant=street_sign_letters

Prompts:
\small
\begin{itemize}[itemsep=1pt,topsep=1pt,leftmargin=12pt]
	\item \texttt{a street sign with the number \cls~}
\end{itemize}
\normalsize &
Class names:

\small
\begin{enumerate}[itemsep=1pt,topsep=1pt,leftmargin=12pt]
	\item \clsfmt{zero}
	\item \clsfmt{one}
	\item \clsfmt{two}
	\item \clsfmt{three}
	\item \clsfmt{four}
	\item \clsfmt{five}
	\item \clsfmt{six}
	\item \clsfmt{seven}
	\item \clsfmt{eight}
	\item \clsfmt{nine}
\end{enumerate}
\normalsize
&
Delta: \small 
+2.4\%
\normalsize \\

\arrayrulecolor{black!30}\midrule
% svhn variant=svhn_heavy

Prompts:
\small
\begin{itemize}[itemsep=1pt,topsep=1pt,leftmargin=12pt]
	\item \texttt{a photo of the number \cls~ written on a sign}
	\item \texttt{an outdoor house number \cls~}
	\item \texttt{the number \cls~ in the center of the image}
	\item \texttt{an outdoor number \cls~ written on a sign}
	\item \texttt{an outdoor number \cls~}
	\item \texttt{a centered image of the number \cls~}
\end{itemize}
\normalsize &
Class names:

\small
\begin{enumerate}[itemsep=1pt,topsep=1pt,leftmargin=12pt]
	\item \clsfmt{0} $\cdot$ \clsfmt{zero}
	\item \clsfmt{1} $\cdot$ \clsfmt{one}
	\item \clsfmt{2} $\cdot$ \clsfmt{two}
	\item \clsfmt{3} $\cdot$ \clsfmt{three}
	\item \clsfmt{4} $\cdot$ \clsfmt{four}
	\item \clsfmt{5} $\cdot$ \clsfmt{five}
	\item \clsfmt{6} $\cdot$ \clsfmt{six}
	\item \clsfmt{7} $\cdot$ \clsfmt{seven}
	\item \clsfmt{8} $\cdot$ \clsfmt{eight}
	\item \clsfmt{9} $\cdot$ \clsfmt{nine}
\end{enumerate}
\normalsize
&
Delta: \small 
+3.2\%
\normalsize \\

\arrayrulecolor{black!100}\bottomrule
\pagebreak
\arrayrulecolor{black!100}\bottomrule
\multicolumn{3}{c}{\textbf{camelyon} \emph{v2.0.0}} \\
% camelyon variant=pathology

\arrayrulecolor{black!100}\midrule
% camelyon variant=pathology_renamed

Prompts:
\small
\begin{itemize}[itemsep=1pt,topsep=1pt,leftmargin=12pt]
	\item \texttt{a histopathology slide showing \cls~}
\end{itemize}
\normalsize &
Class names:

\small
\begin{enumerate}[itemsep=1pt,topsep=1pt,leftmargin=12pt]
	\item \clsfmt{healthy lymph node tissue}
	\item \clsfmt{a lymph node tumor}
\end{enumerate}
\normalsize
&
Delta: \small 
+1.9\%
\normalsize \\

\arrayrulecolor{black!30}\midrule
% camelyon variant=pathology2_renamed

Prompts:
\small
\begin{itemize}[itemsep=1pt,topsep=1pt,leftmargin=12pt]
	\item \texttt{histopathology image of \cls~}
\end{itemize}
\normalsize &
Class names:

\small
\begin{enumerate}[itemsep=1pt,topsep=1pt,leftmargin=12pt]
	\item \clsfmt{healthy lymph node tissue}
	\item \clsfmt{a lymph node tumor}
\end{enumerate}
\normalsize
&
Delta: \small 
+1.9\%
\normalsize \\

\midrule
% camelyon variant=pcam_heavy

Prompts:
\small
\begin{itemize}[itemsep=1pt,topsep=1pt,leftmargin=12pt]
	\item \texttt{an example of \cls~}
	\item \texttt{a histopathology slide of \cls~}
	\item \texttt{an example histopathological image showing \cls~}
	\item \texttt{a histopathology slide showing \cls~}
	\item \texttt{patient's pathology examination indicates \cls~}
	\item \texttt{a \cls~ slide}
\end{itemize}
\normalsize &
Class names:

\small
\begin{enumerate}[itemsep=1pt,topsep=1pt,leftmargin=12pt]
	\item \clsfmt{ healthy tissue} $\cdot$ \clsfmt{  tissue }
	\item \clsfmt{ dangerous tissue} $\cdot$ \clsfmt{  unhealthy tissue }
\end{enumerate}
\normalsize
&
Delta: \small 
+0.8\%
\normalsize \\

\arrayrulecolor{black!100}\bottomrule
\multicolumn{3}{c}{\textbf{eurosat} \emph{v2.0.0}} \\
% eurosat variant=eurosat_heavy
\arrayrulecolor{black!100}\midrule

Prompts:
\small
\begin{itemize}[itemsep=1pt,topsep=1pt,leftmargin=12pt]
	\item \texttt{an overhead view of \cls~}
	\item \texttt{an aerial view of \cls~}
	\item \texttt{an overhead image of \cls~}
	\item \texttt{a satellite photo of \cls~}
	\item \texttt{a satellite image of \cls~}
	\item \texttt{photo of \cls~ from the sky}
\end{itemize}
\normalsize &
Class names:

\small
\begin{enumerate}[itemsep=1pt,topsep=1pt,leftmargin=12pt]
	\item \clsfmt{farmland} $\cdot$ \clsfmt{  farms} $\cdot$ \clsfmt{  an annual crop}
	\item \clsfmt{a forest} $\cdot$ \clsfmt{  woodland} $\cdot$ \clsfmt{  trees}
	\item \clsfmt{a meadow} $\cdot$ \clsfmt{  herbaceous vegetation} $\cdot$ \clsfmt{  grass} $\cdot$ \clsfmt{  fields}
	\item \clsfmt{highway or road} $\cdot$ \clsfmt{  motorways} $\cdot$ \clsfmt{  highways} $\cdot$ \clsfmt{  a street} $\cdot$ \clsfmt{  roads}
	\item \clsfmt{an urban area} $\cdot$ \clsfmt{  an industrial area} $\cdot$ \clsfmt{  an industrial zone} $\cdot$ \clsfmt{  a city} $\cdot$ \clsfmt{  factories}
	\item \clsfmt{a pasture} $\cdot$ \clsfmt{  farmland} $\cdot$ \clsfmt{  farms}
	\item \clsfmt{permanent crop} $\cdot$ \clsfmt{  arable land} $\cdot$ \clsfmt{  an orchard}
	\item \clsfmt{a suburban area} $\cdot$ \clsfmt{  a cul de sac} $\cdot$ \clsfmt{  a residential area} $\cdot$ \clsfmt{  houses}
	\item \clsfmt{a canal} $\cdot$ \clsfmt{  a river} $\cdot$ \clsfmt{  a waterway} $\cdot$ \clsfmt{  a stream}
	\item \clsfmt{an ocean} $\cdot$ \clsfmt{  a water} $\cdot$ \clsfmt{  a sea} $\cdot$ \clsfmt{  a reservoir}
\end{enumerate}
\normalsize
&
Delta: \small 
+6.7\%
\normalsize \\

\arrayrulecolor{black!100}\bottomrule
\pagebreak
\bottomrule
\multicolumn{3}{c}{\textbf{resisc} \emph{v3.0.0}} \\
% resisc variant=resisc_custom
\arrayrulecolor{black!100}\midrule

Prompts:
\small
\begin{itemize}[itemsep=1pt,topsep=1pt,leftmargin=12pt]
	\item \texttt{a satellite image of \cls~}
	\item \texttt{an aerial view of \cls~}
	\item \texttt{a satellite photo of \cls~}
	\item \texttt{\cls~ from above}
\end{itemize}
\normalsize &
Class names:

\small
\begin{enumerate}[itemsep=1pt,topsep=1pt,leftmargin=12pt]
	\item \clsfmt{an airplane} $\cdot$ \clsfmt{  a plane} $\cdot$ \clsfmt{  a flying plane}
	\item \clsfmt{an airfield} $\cdot$ \clsfmt{  an airport} $\cdot$ \clsfmt{  an aeroport}
	\item \clsfmt{baseball diamond} $\cdot$ \clsfmt{  baseball court} $\cdot$ \clsfmt{  baseball} $\cdot$ \clsfmt{  baseball field}
	\item \clsfmt{basketball} $\cdot$ \clsfmt{  a basketball court} $\cdot$ \clsfmt{  an outdoor basketball court}
	\item \clsfmt{beach} $\cdot$ \clsfmt{  sand}
	\item \clsfmt{a walkway} $\cdot$ \clsfmt{  a bridge} $\cdot$ \clsfmt{  a footbridge}
	\item \clsfmt{shrubland} $\cdot$ \clsfmt{  chaparral} $\cdot$ \clsfmt{  sparse plants} $\cdot$ \clsfmt{  desert plants} $\cdot$ \clsfmt{  shrubs} $\cdot$ \clsfmt{  dry plants}
	\item \clsfmt{a church} $\cdot$ \clsfmt{  a chapel}
	\item \clsfmt{circular farmland} $\cdot$ \clsfmt{  circle farm}
	\item \clsfmt{cloudy sky} $\cdot$ \clsfmt{  clouds} $\cdot$ \clsfmt{  cloud}
	\item \clsfmt{a commercial area} $\cdot$ \clsfmt{  a shopping mall} $\cdot$ \clsfmt{  high street} $\cdot$ \clsfmt{  shops}
	\item \clsfmt{densely populated area} $\cdot$ \clsfmt{  lots of houses} $\cdot$ \clsfmt{  a dense residential area} $\cdot$ \clsfmt{ urban area}
	\item \clsfmt{a desert} $\cdot$ \clsfmt{  barren land} $\cdot$ \clsfmt{  sand dunes} $\cdot$ \clsfmt{  wasteland}
	\item \clsfmt{woods} $\cdot$ \clsfmt{  forest} $\cdot$ \clsfmt{  woodland}
	\item \clsfmt{expressway} $\cdot$ \clsfmt{  roads} $\cdot$ \clsfmt{  highway} $\cdot$ \clsfmt{  freeway}
	\item \clsfmt{golf fields} $\cdot$ \clsfmt{  a golf course}
	\item \clsfmt{a running court} $\cdot$ \clsfmt{  a track court} $\cdot$ \clsfmt{  a ground track field}
	\item \clsfmt{a harbor} $\cdot$ \clsfmt{  a dockyard} $\cdot$ \clsfmt{  a haven} $\cdot$ \clsfmt{  a jetty} $\cdot$ \clsfmt{  a quay} $\cdot$ \clsfmt{  a pier}
	\item \clsfmt{an industrial zone} $\cdot$ \clsfmt{  an industrial area} $\cdot$ \clsfmt{  industry}
	\item \clsfmt{a busy intersection} $\cdot$ \clsfmt{  a crash on an intersection} $\cdot$ \clsfmt{  intersection pileup} $\cdot$ \clsfmt{ an intersection}
	\item \clsfmt{an island in the ocean} $\cdot$ \clsfmt{  an island} $\cdot$ \clsfmt{  land surrounded by water} $\cdot$ \clsfmt{  an ocean island}
	\item \clsfmt{a reservoir} $\cdot$ \clsfmt{  a lake} $\cdot$ \clsfmt{  the ocean} $\cdot$ \clsfmt{  the sea}
	\item \clsfmt{a pasture} $\cdot$ \clsfmt{  a paddock} $\cdot$ \clsfmt{  fields} $\cdot$ \clsfmt{  grassland}
	\item \clsfmt{a medium residential area} $\cdot$ \clsfmt{  cul de sac} $\cdot$ \clsfmt{  suburban area} $\cdot$ \clsfmt{  town}
	\item \clsfmt{a mobile home park} $\cdot$ \clsfmt{  caravans} $\cdot$ \clsfmt{  caravan park}
	\item \clsfmt{a mountain} $\cdot$ \clsfmt{  a mountaintop} $\cdot$ \clsfmt{  a hill} $\cdot$ \clsfmt{  a mountain range}
	\item \clsfmt{an overpass}
	\item \clsfmt{a palace} $\cdot$ \clsfmt{  a royal palace} $\cdot$ \clsfmt{  a cheateau}
	\item \clsfmt{parking} $\cdot$ \clsfmt{  a parking lot}
	\item \clsfmt{a train track} $\cdot$ \clsfmt{  a train} $\cdot$ \clsfmt{  a trainline} $\cdot$ \clsfmt{  a rail track} $\cdot$ \clsfmt{  a railway}
	\item \clsfmt{a railway station} $\cdot$ \clsfmt{  a train station}
	\item \clsfmt{rectangular farmland} $\cdot$ \clsfmt{  rectangle farms}
	\item \clsfmt{a river} $\cdot$ \clsfmt{  a stream}
	\item \clsfmt{a roundabout}
	\item \clsfmt{runway} $\cdot$ \clsfmt{  an airport runway} $\cdot$ \clsfmt{  a landing strip}
	\item \clsfmt{an iceberg} $\cdot$ \clsfmt{  ocean ice} $\cdot$ \clsfmt{  sea ice}
	\item \clsfmt{a ship} $\cdot$ \clsfmt{  a boat}
	\item \clsfmt{a snowberg}
	\item \clsfmt{sparsely populated area}
	\item \clsfmt{a stadium} $\cdot$ \clsfmt{  an arena} $\cdot$ \clsfmt{  a football stadium} $\cdot$ \clsfmt{  a sports arena}
	\item \clsfmt{a storage tank} $\cdot$ \clsfmt{  tank}
	\item \clsfmt{a tennis court} $\cdot$ \clsfmt{  tennis} $\cdot$ \clsfmt{  a court} $\cdot$ \clsfmt{  a badminton court}
	\item \clsfmt{rural land} $\cdot$ \clsfmt{  a terrace}
	\item \clsfmt{a power station} $\cdot$ \clsfmt{  a thermal power station}
	\item \clsfmt{a marsh} $\cdot$ \clsfmt{  wetland} $\cdot$ \clsfmt{  peatland} $\cdot$ \clsfmt{  a bog}
\end{enumerate}
\normalsize
&
Delta: \small 
+5.1\%
\normalsize \\

\arrayrulecolor{black!100}\bottomrule
\pagebreak
\bottomrule
\multicolumn{3}{c}{\textbf{clevr-closest} \emph{v3.1.0}} \\
% clevr-closest variant=object_size
\arrayrulecolor{black!100}\midrule

Prompts:
\small
\begin{itemize}[itemsep=1pt,topsep=1pt,leftmargin=12pt]
	\item \texttt{\cls~ objects}
\end{itemize}
\normalsize &
Class names:

\small
\begin{enumerate}[itemsep=1pt,topsep=1pt,leftmargin=12pt]
	\item \clsfmt{massive}
	\item \clsfmt{very large}
	\item \clsfmt{large}
	\item \clsfmt{}
	\item \clsfmt{small}
	\item \clsfmt{very small}
\end{enumerate}
\normalsize
&
Delta: \small 
+0.3\%
\normalsize \\

\arrayrulecolor{black!30}\midrule
% clevr-closest variant=object_distance

Prompts:
\small
\begin{itemize}[itemsep=1pt,topsep=1pt,leftmargin=12pt]
	\item \texttt{\cls~ objects}
\end{itemize}
\normalsize &
Class names:

\small
\begin{enumerate}[itemsep=1pt,topsep=1pt,leftmargin=12pt]
	\item \clsfmt{very nearby}
	\item \clsfmt{nearby}
	\item \clsfmt{near}
	\item \clsfmt{}
	\item \clsfmt{distant}
	\item \clsfmt{very distant}
\end{enumerate}
\normalsize
&
Delta: \small 
+2.7\%
\normalsize \\

\midrule
% clevr-closest variant=shape_size

Prompts:
\small
\begin{itemize}[itemsep=1pt,topsep=1pt,leftmargin=12pt]
	\item \texttt{\cls~ shapes}
\end{itemize}
\normalsize &
Class names:

\small
\begin{enumerate}[itemsep=1pt,topsep=1pt,leftmargin=12pt]
	\item \clsfmt{massive}
	\item \clsfmt{very large}
	\item \clsfmt{large}
	\item \clsfmt{}
	\item \clsfmt{small}
	\item \clsfmt{very small}
\end{enumerate}
\normalsize
&
Delta: \small 
+0.6\%
\normalsize \\

\midrule
% clevr-closest variant=shape_distance

Prompts:
\small
\begin{itemize}[itemsep=1pt,topsep=1pt,leftmargin=12pt]
	\item \texttt{\cls~ shapes}
\end{itemize}
\normalsize &
Class names:

\small
\begin{enumerate}[itemsep=1pt,topsep=1pt,leftmargin=12pt]
	\item \clsfmt{very nearby}
	\item \clsfmt{nearby}
	\item \clsfmt{near}
	\item \clsfmt{}
	\item \clsfmt{distant}
	\item \clsfmt{very distant}
\end{enumerate}
\normalsize
&
Delta: \small 
+3.9\%
\normalsize \\

\midrule
% clevr-closest variant=closest_heavy

Prompts:
\small
\begin{itemize}[itemsep=1pt,topsep=1pt,leftmargin=12pt]
	\item \texttt{\cls~ thing}
	\item \texttt{the nearest shape in this image is \cls~}
	\item \texttt{the closest shape in this rendered image is \cls~}
	\item \texttt{the closest shape in this image is \cls~}
\end{itemize}
\normalsize &
Class names:

\small
\begin{enumerate}[itemsep=1pt,topsep=1pt,leftmargin=12pt]
	\item \clsfmt{ huge} $\cdot$ \clsfmt{  super near }
	\item \clsfmt{ nearby }
	\item \clsfmt{ big} $\cdot$ \clsfmt{  large }
	\item \clsfmt{ quite small} $\cdot$ \clsfmt{  medium sized} $\cdot$ \clsfmt{  normal sized }
	\item \clsfmt{ small} $\cdot$ \clsfmt{  distant }
	\item \clsfmt{ very small} $\cdot$ \clsfmt{  very distant }
\end{enumerate}
\normalsize
&
Delta: \small 
+1.8\%
\normalsize \\

\arrayrulecolor{black!100}\bottomrule
\pagebreak
\arrayrulecolor{black!100}\bottomrule
\multicolumn{3}{c}{\textbf{clevr-count} \emph{v3.1.0}} \\
% clevr-count variant=count1
\midrule
% clevr-count variant=count1_letters

Prompts:
\small
\begin{itemize}[itemsep=1pt,topsep=1pt,leftmargin=12pt]
	\item \texttt{\cls~ objects}
\end{itemize}
\normalsize &
Class names:

\small
\begin{enumerate}[itemsep=1pt,topsep=1pt,leftmargin=12pt]
	\item \clsfmt{three}
	\item \clsfmt{four}
	\item \clsfmt{five}
	\item \clsfmt{six}
	\item \clsfmt{seven}
	\item \clsfmt{eight}
	\item \clsfmt{nine}
	\item \clsfmt{ten}
\end{enumerate}
\normalsize
&
Delta: \small 
+0.4\%
\normalsize \\

\arrayrulecolor{black!30}\midrule
% clevr-count variant=count2_letters

Prompts:
\small
\begin{itemize}[itemsep=1pt,topsep=1pt,leftmargin=12pt]
	\item \texttt{\cls~ things}
\end{itemize}
\normalsize &
Class names:

\small
\begin{enumerate}[itemsep=1pt,topsep=1pt,leftmargin=12pt]
	\item \clsfmt{three}
	\item \clsfmt{four}
	\item \clsfmt{five}
	\item \clsfmt{six}
	\item \clsfmt{seven}
	\item \clsfmt{eight}
	\item \clsfmt{nine}
	\item \clsfmt{ten}
\end{enumerate}
\normalsize
&
Delta: \small 
+0.6\%
\normalsize \\

\midrule
% clevr-count variant=count3_letters

Prompts:
\small
\begin{itemize}[itemsep=1pt,topsep=1pt,leftmargin=12pt]
	\item \texttt{a photo of \cls~ objects}
\end{itemize}
\normalsize &
Class names:

\small
\begin{enumerate}[itemsep=1pt,topsep=1pt,leftmargin=12pt]
	\item \clsfmt{three}
	\item \clsfmt{four}
	\item \clsfmt{five}
	\item \clsfmt{six}
	\item \clsfmt{seven}
	\item \clsfmt{eight}
	\item \clsfmt{nine}
	\item \clsfmt{ten}
\end{enumerate}
\normalsize
&
Delta: \small 
+0.7\%
\normalsize \\

\midrule
% clevr-count variant=counting_heavy

Prompts:
\small
\begin{itemize}[itemsep=1pt,topsep=1pt,leftmargin=12pt]
	\item \texttt{a picture of \cls~}
	\item \texttt{there are \cls~}
	\item \texttt{there are \cls~ in the image}
	\item \texttt{a rendered image of \cls~}
\end{itemize}
\normalsize &
Class names:

\small
\begin{enumerate}[itemsep=1pt,topsep=1pt,leftmargin=12pt]
	\item \clsfmt{ 3 objects} $\cdot$ \clsfmt{ three objects} $\cdot$ \clsfmt{ 3 shapes} $\cdot$ \clsfmt{ three shapes }
	\item \clsfmt{ 4 objects} $\cdot$ \clsfmt{ four objects} $\cdot$ \clsfmt{ 4 shapes} $\cdot$ \clsfmt{ four shapes }
	\item \clsfmt{ 5 objects} $\cdot$ \clsfmt{ five objects} $\cdot$ \clsfmt{ 5 shapes} $\cdot$ \clsfmt{ five shapes }
	\item \clsfmt{ 6 objects} $\cdot$ \clsfmt{ six objects} $\cdot$ \clsfmt{ 6 shapes} $\cdot$ \clsfmt{ six shapes }
	\item \clsfmt{ 7 objects} $\cdot$ \clsfmt{ seven objects} $\cdot$ \clsfmt{ 7 shapes} $\cdot$ \clsfmt{ seven shapes }
	\item \clsfmt{ 8 objects} $\cdot$ \clsfmt{ eight objects} $\cdot$ \clsfmt{ 8 shapes} $\cdot$ \clsfmt{ eight shapes }
	\item \clsfmt{ 9 objects} $\cdot$ \clsfmt{ nine objects} $\cdot$ \clsfmt{ 9 shapes} $\cdot$ \clsfmt{ nine shapes }
	\item \clsfmt{ 10 objects} $\cdot$ \clsfmt{ ten objects} $\cdot$ \clsfmt{ 10 shapes} $\cdot$ \clsfmt{ ten shapes }
\end{enumerate}
\normalsize
&
Delta: \small 
+1.2\%
\normalsize \\

\arrayrulecolor{black!100}\bottomrule
\\
\caption{\normalsize Prompts and customized class names swept over for VTAB tasks. Performance deltas are shown as mean test accuracy improvement per-task compared to just using the default three prompts.}
\label{tab:vtab_prompts_custom}
\end{longtable}
\twocolumn

\begin{table*}
% \vspace{-1cm}
\centering
\label{table:main}
\footnotesize
\resizebox{\textwidth}{!}{%
\addtolength{\tabcolsep}{-1.7pt}% <- remove some horizontal white space between the columns for better text readability
\begin{tabular}{@{}llrlll@{\hspace{0.3ex}}llllrlrrrrrr@{}}
\toprule
                                      Ref & Dataset & Images &   Cfg &  H &        Image &        Text & Tok &     Inits & Optim &   LR &    WD & INet & T$\rightarrow$I & I$\rightarrow$T &   Vn &  Vsp &  Vst \\
\midrule
              Fig~\ref{fig:teaser_curve} &    YFCC\textsubscript{CLIP} &   983M &  \LU &  y &     vit-B/32 &   bert-base &  WP &   AR,Bert &  Adam & 8e-4 &  1e-4 & 63.6 &            22.1 &            37.6 & 59.3 & 35.0 & 12.7 \\
              Fig~\ref{fig:teaser_curve} &    YFCC\textsubscript{CLIP} &   983M &  \UU &  y &     vit-B/32 &   bert-base &  WP &   AR,Bert &  Adam & 3e-4 &  1e-5 & 53.3 &            23.4 &            37.6 & 54.9 & 44.4 & 14.1 \\
              Fig~\ref{fig:teaser_curve} &    YFCC\textsubscript{CLIP} &   983M &  \uu &  y &     vit-B/32 &   bert-base &  WP &       -,- &  Adam & 8e-4 &  1e-4 & 42.1 &            17.9 &            31.1 & 45.8 & 49.8 & 14.3 \\
    \arrayrulecolor{lightgray}\midrule[0.25pt]\arrayrulecolor{black}
                     Tab~\ref{table:sota} &    Ours &  18.2B &  \Lu &  n &    vit-g/14* &   vit-giant &  SP &     JFT,- &  Adaf & 1e-3 &     0 & 85.2 &            41.9 &            59.3 & 74.7 & - & - \\
                     Tab~\ref{table:sota} &   Mixed &   983M &  \LU &  y &     vit-L/16 &  bert-large &  WP &   AR,Bert &  Adaf & 8e-4 &  1e-4 & 75.7 &            31.2 &            48.5 & 63.1 & 50.3 & 14.1 \\
    \arrayrulecolor{lightgray}\midrule[0.25pt]\arrayrulecolor{black}
              Tab~\ref{table:design_our} &    Ours &   901M &  \Lu &  n &     vit-B/32 &    vit-base &  SP &     JFT,- &  Adaf & 1e-3 &     0 & 70.1 &            28.6 &            43.8 & 66.6 & 57.2 & 14.6 \\
              Tab~\ref{table:design_our} &    Ours &   901M &  \Uu &  y &     vit-B/32 &    vit-base &  SP &     JFT,- &  Adaf & 1e-3 &     0 & 57.2 &            27.0 &            40.1 & 60.1 & 58.0 & 15.0 \\
              Tab~\ref{table:design_our} &    Ours &   901M &  \uu &  y &     vit-B/32 &    vit-base &  SP &       -,- &  Adaf & 1e-3 &     0 & 50.6 &            24.1 &            38.9 & 55.3 & 38.9 & 16.5 \\
    \arrayrulecolor{lightgray}\midrule[0.25pt]\arrayrulecolor{black}
  Tab~\ref{table:more_image_pretrainings} &    YFCC\textsubscript{CLIP} &   246M &  \LU &  y &    dino-B/16 &   bert-base &  WP &  vit,Bert &  Adam & 8e-4 &  1e-4 & 55.5 &            18.2 &            33.4 & 51.5 & 45.4 & 14.8 \\
  Tab~\ref{table:more_image_pretrainings} &    YFCC\textsubscript{CLIP} &   246M &  \LU &  y &  mocov3-B/16 &   bert-base &  WP &  vit,Bert &  Adam & 8e-4 &  1e-4 & 55.4 &            17.6 &            33.5 & 50.8 & 40.5 & 12.8 \\
    \arrayrulecolor{lightgray}\midrule[0.25pt]\arrayrulecolor{black}
 Tab~\ref{table:more_image_architectures} &   CC12M &   200M &  \LU &  n &     vit-B/32 &   bert-base &  WP &   AR,Bert &  Adam & 1e-3 &  1e-4 & 60.7 &            25.0 &            41.3 & 57.7 & 49.6 & 13.9 \\
 Tab~\ref{table:more_image_architectures} &   CC12M &   200M &  \LU &  n &     bit-50x1 &   bert-base &  WP &    M,Bert &  Adam & 1e-3 &  1e-4 & 55.2 &            23.9 &            37.3 & 53.2 & 49.3 & 14.3 \\
 Tab~\ref{table:more_image_architectures} &   CC12M &   200M &  \LU &  n &   mixer-B/32 &   bert-base &  WP &   AR,Bert &  Adam & 1e-3 &  1e-4 & 57.1 &            22.9 &            37.5 &    - &    - &    - \\
    \arrayrulecolor{lightgray}\midrule[0.25pt]\arrayrulecolor{black}
              Tab~\ref{tab:text_encoder} &    YFCC &   901M &  \LU &  y &     vit-B/32 &    mt5-base &  SP &    AR,mt5 &  Adam & 8e-4 &  1e-4 & 59.3 &            17.4 &            28.7 & 55.5 & 47.3 & 15.2 \\
              Tab~\ref{tab:text_encoder} &    YFCC &   901M &  \Lu &  y &     vit-B/32 &    vit-base &  WP &      AR,- &  Adam & 8e-4 &  1e-4 & 56.4 &            17.3 &            28.2 & 53.3 & 47.4 & 14.1 \\
              Tab~\ref{tab:text_encoder} &    YFCC &   901M &  \LU &  y &     vit-B/32 &   bert-base &  WP &   AR,Bert &  Adam & 8e-4 &  1e-4 & 59.5 &            20.7 &            36.3 & 56.7 & 51.3 & 12.3 \\
              Tab~\ref{tab:text_encoder} &    YFCC &   901M &  \Lu &  y &     vit-B/32 &    mt5-base &  SP &      AR,- &  Adam & 8e-4 &  1e-4 & 58.1 &            16.4 &            28.3 & 54.7 & 41.8 & 14.4 \\
              Tab~\ref{tab:text_encoder} &    YFCC &   901M &  \Lu &  y &     vit-B/32 &   bert-base &  WP &      AR,- &  Adam & 8e-4 &  1e-4 & 58.8 &            20.0 &            35.2 & 55.2 & 51.8 & 14.6 \\
              Tab~\ref{tab:text_encoder} &    YFCC &   901M &  \Lu &  y &     vit-B/32 &    vit-base &  SP &      AR,- &  Adam & 1e-3 &  1e-4 & 57.2 &            16.9 &            29.7 & 54.6 & 47.4 & 13.5 \\
              Tab~\ref{tab:text_encoder} &    YFCC &   901M &  \LU &  y &     vit-B/32 &     t5-base &  SP &     AR,t5 &  Adam & 1e-3 &  1e-4 & 59.2 &            18.4 &            31.0 & 57.1 & 47.6 & 14.1 \\
              Tab~\ref{tab:text_encoder} &    YFCC &   901M &  \Lu &  y &     vit-B/32 &     t5-base &  SP &      AR,- &  Adam & 1e-3 &  1e-4 & 57.8 &            17.2 &            29.4 & 54.5 & 46.3 & 13.2 \\
    \arrayrulecolor{lightgray}\midrule[0.25pt]\arrayrulecolor{black}
      Fig~\ref{fig:model_capacity_impact} &   CC12M &   200M &  \LU &  n &     vit-B/16 &  bert-large &  WP &   AR,Bert &  Adaf & 1e-3 &  1e-4 & 66.9 &            28.3 &            44.8 & 58.6 & 45.4 & 13.5 \\
      Fig~\ref{fig:model_capacity_impact} &   CC12M &   200M &  \LU &  n &     vit-L/16 &  bert-large &  WP &   AR,Bert &  Adaf & 1e-3 &  1e-4 & 67.6 &            26.9 &            42.6 & 57.8 & 50.3 & 13.0 \\
      Fig~\ref{fig:model_capacity_impact} &   CC12M &   200M &  \LU &  n &     vit-B/16 &   bert-base &  WP &   AR,Bert &  Adam & 1e-3 &  1e-4 & 66.1 &            28.2 &            45.3 & 59.0 & 50.6 & 14.0 \\
      Fig~\ref{fig:model_capacity_impact} &   CC12M &   200M &  \LU &  n &     vit-L/16 &   bert-base &  WP &   AR,Bert &  Adam & 1e-3 &  1e-4 & 66.8 &            26.6 &            44.3 & 58.6 & 45.6 & 12.7 \\
      Fig~\ref{fig:model_capacity_impact} &   CC12M &   200M &  \LU &  n &     vit-B/32 &  bert-large &  WP &   AR,Bert &  Adaf & 1e-3 &  1e-4 & 61.7 &            25.4 &            41.4 & 56.4 & 49.9 & 13.6 \\
      Fig~\ref{fig:model_capacity_impact} &   CC12M &   200M &  \LU &  n &     vit-B/32 &   bert-base &  WP &   AR,Bert &  Adam & 1e-3 &  1e-4 & 61.1 &            24.9 &            40.9 & 56.8 & 49.6 & 15.4 \\
    \arrayrulecolor{lightgray}\midrule[0.25pt]\arrayrulecolor{black}
      Fig~\ref{fig:model_capacity_impact} &    Ours &   901M &  \Lu &  n &     vit-g/14 &    vit-huge &  SP &     JFT,- &  Adaf & 1e-3 &     0 & 81.8 &            33.1 &            48.9 & 70.6 & 61.4 & 15.2 \\
      Fig~\ref{fig:model_capacity_impact} &    Ours &   901M &  \Lu &  n &     vit-g/14 &   vit-large &  SP &     JFT,- &  Adaf & 1e-3 &     0 & 81.2 &            32.9 &            48.5 & 69.2 & 50.5 & 15.3 \\
      Fig~\ref{fig:model_capacity_impact} &    Ours &   901M &  \Lu &  n &     vit-L/16 &    vit-huge &  SP &     JFT,- &  Adaf & 1e-3 &     0 & 80.8 &            35.6 &            51.2 & 69.2 & 50.3 & 13.5 \\
      Fig~\ref{fig:model_capacity_impact} &    Ours &   901M &  \Lu &  n &     vit-L/16 &   vit-large &  SP &     JFT,- &  Adaf & 1e-3 &     0 & 80.3 &            34.8 &            49.8 & 68.9 & 60.3 & 14.8 \\
      Fig~\ref{fig:model_capacity_impact} &    Ours &   901M &  \Lu &  n &     vit-g/14 &    vit-base &  SP &     JFT,- &  Adaf & 1e-3 &     0 & 79.5 &            30.7 &            45.9 & 68.6 & 59.6 & 12.6 \\
      Fig~\ref{fig:model_capacity_impact} &    Ours &   901M &  \Lu &  n &     vit-B/16 &    vit-huge &  SP &     JFT,- &  Adaf & 1e-3 &     0 & 77.1 &            34.5 &            49.7 & 68.0 & 59.7 & 14.0 \\
      Fig~\ref{fig:model_capacity_impact} &    Ours &   901M &  \Lu &  n &     vit-L/16 &    vit-base &  SP &     JFT,- &  Adaf & 1e-3 &     0 & 78.5 &            33.5 &            48.6 & 68.2 & 61.0 & 13.8 \\
      Fig~\ref{fig:model_capacity_impact} &    Ours &   901M &  \Lu &  n &     vit-B/16 &   vit-large &  SP &     JFT,- &  Adaf & 1e-3 &     0 & 76.8 &            33.6 &            49.4 & 68.5 & 45.0 & 14.2 \\
      Fig~\ref{fig:model_capacity_impact} &    Ours &   901M &  \Lu &  n &     vit-B/16 &    vit-base &  SP &     JFT,- &  Adaf & 1e-3 &     0 & 75.2 &            31.9 &            46.8 & 67.5 & 57.7 & 12.8 \\
      Fig~\ref{fig:model_capacity_impact} &    Ours &   901M &  \Lu &  n &     vit-B/32 &    vit-huge &  SP &     JFT,- &  Adaf & 1e-3 &     0 & 72.2 &            31.2 &            46.4 & 68.3 & 55.1 & 13.8 \\
      Fig~\ref{fig:model_capacity_impact} &    Ours &   901M &  \Lu &  n &     vit-B/32 &   vit-large &  SP &     JFT,- &  Adaf & 1e-3 &     0 & 71.6 &            30.7 &            45.6 & 66.4 & 55.0 & 14.5 \\
      Fig~\ref{fig:model_capacity_impact} &    Ours &   901M &  \Lu &  n &     vit-B/32 &    vit-base &  SP &     JFT,- &  Adaf & 1e-3 &     0 & 70.0 &            29.2 &            43.8 & 65.8 & 56.9 & 12.0 \\
    \arrayrulecolor{lightgray}\midrule[0.25pt]\arrayrulecolor{black}
                  Fig~\ref{fig:xlang_i1k} &    YFCC\textsubscript{CLIP} &   983M &  \LU &  y &     vit-B/32 &    mt5-base &  SP &    AR,mt5 &  Adam & 3e-4 &  1e-4 & 58.4 &            15.6 &            25.1 & 54.5 & 36.7 & 12.3 \\
                  Fig~\ref{fig:xlang_i1k} &    YFCC\textsubscript{CLIP} &   983M &  \LU &  y &     vit-B/32 &     t5-base &  SP &     AR,t5 &  Adam & 3e-4 &  1e-4 & 58.5 &            17.2 &            29.1 & 54.7 & 40.4 & 13.6 \\
                  Fig~\ref{fig:xlang_i1k} &    YFCC\textsubscript{CLIP} &   983M &  \Lu &  y &     vit-B/32 &    mt5-base &  SP &      AR,- &  Adam & 1e-3 &  1e-5 & 58.7 &            14.4 &            23.1 & 53.1 & 41.3 & 14.7 \\
                  Fig~\ref{fig:xlang_i1k} &    YFCC\textsubscript{CLIP} &   983M &  \Lu &  y &     vit-B/32 &     t5-base &  SP &      AR,- &  Adam & 8e-4 &  1e-4 & 58.9 &            14.5 &            22.6 & 53.1 & 41.6 & 15.0 \\
    \arrayrulecolor{lightgray}\midrule[0.25pt]\arrayrulecolor{black}
                  Fig~\ref{fig:xlang_i1k} &    YFCC &   983M &  \LU &  y &     vit-B/32 &    mt5-base &  SP &    AR,mt5 &  Adam & 8e-4 &  1e-4 & 62.6 &            18.9 &            33.6 & 59.0 & 47.6 & 13.8 \\
                  Fig~\ref{fig:xlang_i1k} &    YFCC &   983M &  \Lu &  y &     vit-B/32 &    mt5-base &  SP &      AR,- &  Adam & 8e-4 &  1e-4 & 62.1 &            18.5 &            32.6 & 58.7 & 50.0 & 14.8 \\
                  Fig~\ref{fig:xlang_i1k} &    YFCC &   983M &  \Lu &  y &     vit-B/32 &     t5-base &  SP &      AR,- &  Adam & 1e-3 &  1e-4 & 62.4 &            19.6 &            34.3 & 60.8 & 31.5 & 14.8 \\
                  Fig~\ref{fig:xlang_i1k} &    YFCC &   983M &  \LU &  y &     vit-B/32 &     t5-base &  SP &     AR,t5 &  Adam & 1e-3 &  1e-4 & 62.3 &            20.1 &            34.5 & 61.1 & 50.3 & 14.6 \\
\bottomrule
\end{tabular}%
}% closing brae for resizebox

\caption{Detailed configuration and metrics for a selection of models. 
\emph{Ref} describes the Figure/Table where the model is mentioned. 
\emph{Dataset} describes the dataset that was used (see Section~\ref{sec:data}), with “Mixed” referring to alternating batches between CC12M and YFCC100m.
\emph{Images} is the number of images seen during contrastive-tuning. Default batch size was 16\,384 (only exception model “g/14*” with 32\,768).
\emph{Cfg} first letter refers to image tower, second letter to text tower (Section~\ref{sec:design_choices}).
\emph{H} describes whether a linear head was added to the image tower (note that the text tower always has a linear head).
\emph{Image} describes the image tower (all models use 224px input resolution apart from “g/14*” that uses 288px), for details on models see~\cite{vit,vitg,dino,mocov3,bit,mixer}.
\emph{Text} describes the text tower, for details see~\cite{vit,bert,T5,mt5}.
\emph{Tok} describes whether a SentencePiece or WordPiece tokenizer was used.
\emph{Inits} describes the initializations of the image/text towers (AR refers to AugReg “recommended checkpoints” \cite{augreg}).
\emph{Optim} is the optimizer, using default Adam or Adafactor~\cite{adafactor}.
\emph{LR} is the base learning rate (with linear ramp-up and cosine decay).
\emph{WD} is the weight decay (using “decoupled” weight decay~\cite{decoupled}).
\emph{INet} describes zero-shot top-1 accuracy on Imagenet.
\emph{T$\rightarrow$I} and \emph{I$\rightarrow$T} describe retrieval recall @1 on the MSCOCO test set.
\emph{Vn}, \emph{Vsp}, \emph{Vst} VTAB~\cite{vtab} results for “natural”, “specialized”, and “structured” subsets.
}
% “”
\end{table*}

\end{document}